\RequirePackage{fix-cm}
\documentclass[preprint]{elsarticle} 

\usepackage{graphicx}
\usepackage{multirow}
\usepackage{url}
\usepackage{amssymb}
\usepackage{algorithm}
\usepackage{algpseudocode}
\usepackage{natbib} 
\usepackage{minibox}
\usepackage{bm}
\usepackage{float}

\usepackage{siunitx}
\usepackage{booktabs}

\journal{Pattern Recognition}

\usepackage[cmex10]{amsmath}
\usepackage{amssymb}
\usepackage{mathtools}

\newcommand{\ignore}[1]{}

\newcommand{\e}{\ensuremath{\gamma}}

\newcommand{\vect}[1]{\ensuremath{\mathbf{#1}}}

\newcommand{\Prob}{\ensuremath{\mathbf{P}}}

\newcommand{\todo}[1]{{\bf TODO:}#1\\
\line(1,0){200}

}
\renewcommand{\todo}[1]{}

\newcommand{\eqdef}{\ensuremath{\stackrel{\mathrm{def}}{=}}}

\newcommand{\bigO}{\ensuremath{\mathcal{O}}}

\newcommand{\ttxt}[1]{\ensuremath{\textrm{#1}}}
\newcommand{\tvar}[1]{\ensuremath{\textit #1}}

\newcommand{\tcap}{} 
\newcommand{\tsys}[1]{\ensuremath{\mbox{\sc {#1}}}}

\graphicspath{{./svg_images/}}

\algrenewcommand{\algorithmiccomment}[1]{\hfill\% #1}

\begin{document}

\begin{frontmatter}

\title{Improved graph-based SFA: Information preservation \\ complements the slowness principle}

\author{Alberto N. Escalante-B.\corref{cor1}}
\ead{alberto.escalante@ini.rub.de}

\author{Laurenz Wiskott\corref{}}
\ead{laurenz.wiskott@ini.rub.de}
\address{Institut f\"ur Neuroinformatik \\ 
         Ruhr-University Bochum, Germany }  
\cortext[cor1]{Corresponding author.}

\begin{abstract}
Slow feature analysis (SFA) is an unsupervised-learning algorithm that extracts slowly varying features from a multi-dimensional time series.
A supervised extension to SFA for classification and regression is graph-based SFA (GSFA). GSFA is based on the preservation of similarities, which are specified by a graph structure derived from the labels.
It has been shown that hierarchical GSFA (HGSFA) allows learning from images and other high-dimensional data.
%
%
The feature space spanned by HGSFA is complex due to the composition of the nonlinearities of the nodes in the network. 
However, we show that the network discards useful information prematurely before it reaches higher nodes, resulting in suboptimal global slowness and an under-exploited feature space.

To counteract these problems, we propose an extension called hierarchical information-preserving GSFA (HiGSFA), where information preservation complements the slowness-maximization goal.
We build a 10-layer HiGSFA network to estimate human age from facial photographs of the MORPH-II database, achieving a mean absolute error of 3.50 years, improving the state-of-the-art performance.
HiGSFA and HGSFA support multiple-labels and offer a rich feature space, feed-forward training, and linear complexity in the number of samples and dimensions. 
Furthermore, HiGSFA outperforms HGSFA in terms of feature slowness, estimation accuracy and input reconstruction, giving rise to a promising hierarchical supervised-learning approach.  
\end{abstract}


\end{frontmatter}





\section{Introduction}
\label{sec:intro}
Supervised dimensionality reduction (supervised DR) has promising applications in computer vision, pattern recognition and machine learning, where it is frequently used as a pre-processing step to solve classification and regression problems with high-dimensional input.
The goal of supervised DR is to extract a low-dimensional representation of the input samples that contains the predictive information about the labels (e.g., \citep{RisGraCecPerGor-2008}). One advantage is that dimensions irrelevant to the supervised problem can be discarded, resulting in a more compact representation and more accurate estimations.

The slowness principle requires the extraction of the most slowly changing features, see~\citep{Hinton-1989}, and can be used for DR by itself or before other algorithms. 
It has been shown that this principle might explain in part how the neurons in the brain self-organize to compute invariant representations. 
Slow features can be computed using a few methods, such as online learning rules (e.g., \citep{Foldiak-1991, Mitchison-1991}), slow feature analysis (SFA)~\citep{Wiskott-1998a,WisSej2002}, which is a closed-form algorithm specific for this task that has biologically feasible variants~\citep{Sprekeler-EtAl-2007}, and an incremental-learning version (inc-SFA)~\citep{KompellaEtAl-2012}.

The optimization objective of SFA is the minimization of the squared output differences between (temporally) consecutive pairs of samples. Although SFA is unsupervised, the order of the samples can be regarded as a weak form of supervised information, allowing the use of SFA for supervised learning tasks.

Graph-based SFA (GSFA) \cite{EscalanteWiskott-2012a,EscalanteWiskott-2013b} is an extension of SFA explicitly designed for the solution of classification and regression problems and yields more accurate label estimations than SFA.
GSFA has been used to estimate age and gender from synthetic face images~\cite{EscalanteWiskott-2010}, and more recently for traffic sign classification~\cite{EscalanteWiskott-2015a} and face detection~\cite{MohMah2010}.
GSFA is trained with a training graph, where the vertices are the samples and the edges represent similarities of the corresponding labels.
The optimization objective of GSFA is the minimization of weighted squared output differences between samples connected by an edge, where the edges are carefully chosen to reflect label similarities.



\makeatletter%
\if@twocolumn%
	\newcommand{\mysizeversionssfa}{0.90\columnwidth}
\else
	\newcommand{\mysizeversionssfa}{0.75\columnwidth}
\fi
\makeatother
\begin{figure}[hbt!]
\begin{small}
\begin{center}
\includegraphics[width=\mysizeversionssfa]{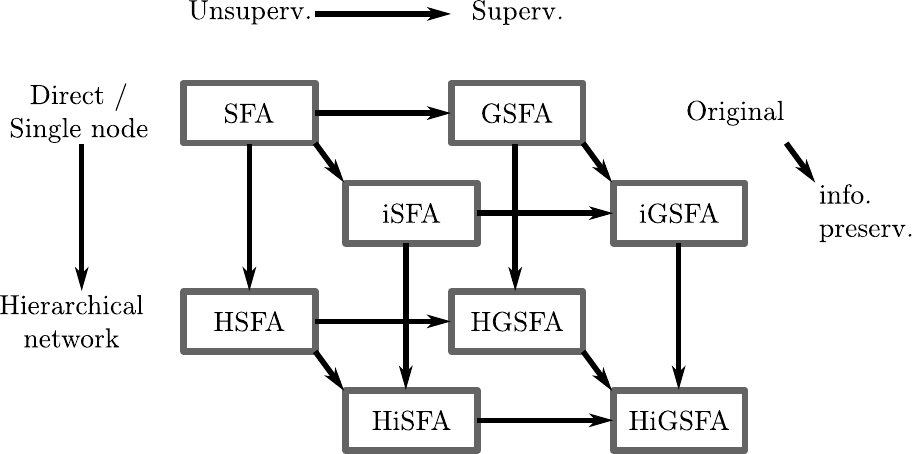}
\caption{The combination of three extensions of SFA (graph-based, hierarchical, information-preserving) gives rise to 8 different versions of SFA. In this article we propose the versions with information preservation, being HiGSFA the most significant one.}
\label{fig:versionsSFA} 
\end{center}
\end{small}
\end{figure}

A common problem in machine learning is the high computational cost of algorithms when the data are high-dimensional.
This is also the case when GSFA is applied to the data directly (direct GSFA), because it has cubic complexity w.r.t.\ the number of dimensions.
However, processing high-dimensional data is still practical if one resorts to hierarchical GSFA (HGSFA).
Instead of extracting slow features in a single step, 
the features are computed by applying GSFA on lower dimensional data chunks repeatedly.
%
Other advantages of HGSFA over direct GSFA include lower memory requirements and a complex global nonlinearity due to the composition of nonlinear transformations through the layers. 
Moreover, the local extraction of slow features from lower-dimensional data chunks typically results in less overfitting. 

In this article, we show that HGSFA suffers from a drawback:
The separate instances of GSFA that process the low-dimensional data chunks in the network, also called GSFA nodes, may prematurely discard features that are not slow at a local level but that would have been useful to improve global slowness (i.e., the slowness of the final output features) if combined with features from other nodes.
This drawback, which we call unnecessary information loss, leads to a suboptimal use of the feature space comprised by the network and also affects the estimation accuracy for the underlying supervised learning problem.

\begin{centering} \begin{table}[hbt!] \begin{center} \centering 
\footnotesize
\begin{tabular}{ll}
\toprule
{\bf Principle or heuristic} & {\bf Implemented through} \\ 
\midrule
Slowness principle, &  \multirow{3}{*}{\parbox{3.5cm}{GSFA, supervised training graphs}}\\ 
exploit label similarities &  \\
through slowness &  \\
\midrule
Spatial localization of features, & \multirow{2}{*}{Hierarchical processing} \\ 
divide and conquer approach &  \\
\midrule
\multirow{2}{*}{Robustness to outliers} & Normalized or saturating \\ 
                                        & nonlinear expansions \\
\midrule
Information preservation  & Minimization of a \\ 
(new)                     & reconstruction error, PCA \\
\midrule
Multiple information  & Multi-label learning combining \\ 
channels (new)        & efficient training graphs \\
\bottomrule
\end{tabular}
\end{center} \caption{Principles, heuristics and ideas considered in HiGSFA and the base methods or algorithms used to exploit them.}
\label{tab:principles} \end{table} \end{centering}

To reduce the unnecessary information loss in HGSFA, we propose to complement slowness with information preservation (i.e., maximization of mutual information between the input data and the output features).
For simplicity and efficiency, we implement this idea as the minimization of a reconstruction error.
The resulting network is called hierarchical information-preserving GSFA (HiGSFA), and the algorithm constituting each node of the network is called information-preserving GSFA (iGSFA).
The features computed by iGSFA can be divided in two parts: A slow part, which is a linear transformation of the (nonlinear) features computed with GSFA, and an input-reconstructive part.
To compute the input-reconstructive part the input data are approximated using the slow part, resulting in residual data that are then processed by PCA.
The construction ensures that both parts are decorrelated and the features have compatible scales. 
Different versions of SFA with and without information preservation are shown in Figure~\ref{fig:versionsSFA}. The principles and heuristics behind HiGSFA are presented in Table~\ref{tab:principles}. 


The experiments show the advantages of HiGSFA over HGSFA: (1) slower features, (2) better generalization to unseen data, (3) much better input reconstruction (see Figure~\ref{fig.input_images}), and (4) improved accuracy for the supervised learning problem.
Furthermore, the computational and memory requirements of HiGSFA are asymptotically the same as in HGSFA.

\makeatletter%
\if@twocolumn%
	\newcommand{\mysizerec}{0.85\columnwidth}
\else
	\newcommand{\mysizerec}{0.6\columnwidth}
\fi
\makeatother
\begin{figure}[hbt!]
\begin{small}
\begin{center}
\includegraphics[width=\mysizerec]{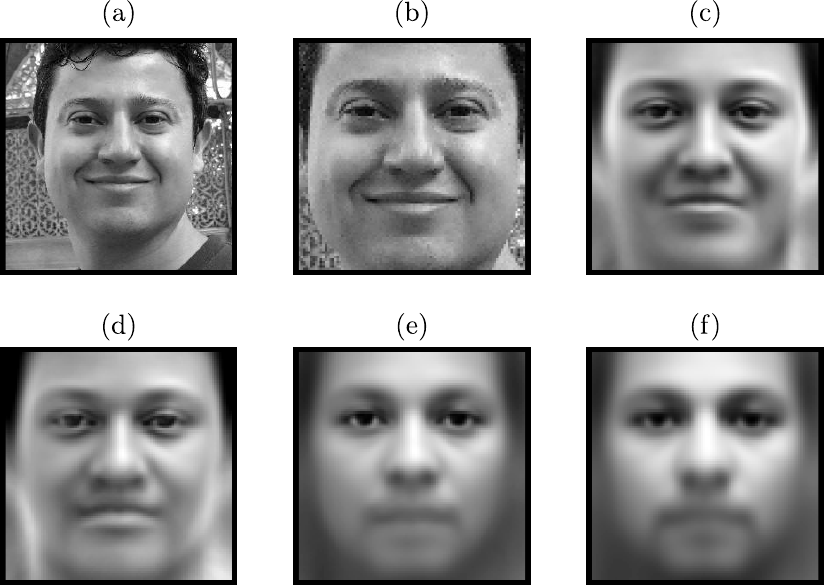}
\caption{
(a) An image from a private database after pose normalization. 
(b) The same image fully pre-processed (i.e., after pose normalization and face sampling, 96$\times$96 pixels).
Linear reconstructions on 75 features extracted with either (c) PCA, (d) HiGSFA or (e) HGSFA.
(f) Average over all pre-processed images of the MORPH-II database. 
}
\label{fig.input_images}
\end{center}
\end{small}
\end{figure}

\section{Related work}
HiGSFA is the main extension to SFA proposed in this work and belongs to supervised DR. Other algorithms for supervised DR include Fisher discriminant analysis (FDA) \citep{Fisher36}, local FDA (LFDA) \citep{Sugiyama-2006}, pairwise constraints-guided feature projection (PCGFP) \citep{TanZho-2007}, semi-supervised dimensionality reduction (SSDR) \citep{ZhaZhoChe-2007}, and semi-supervised LFDA (SELF) \citep{SugIdeNakSes-2010}. 

Existing extensions to SFA include extended SFA (xSFA) \citep{SprekelerZitoEtAl-2010}, generalized SFA (genSFA)~\citep{Sprekeler-2011} and graph-based SFA (GSFA)~\citep{EscalanteWiskott-2012a,EscalanteWiskott-2013b,EscalanteWiskott-2010}.
HiGSFA extends hierarchical GSFA (HGSFA) by adding information preservation.
With some adaptations, SFA has been shown to be successful for classification (e.g., \citep{FWW08, Berkes-2005a,KochKonenHein-2010,KuhnlKummertEtAl-2011,ZhangTao-2012}), and regression (e.g., \citep{FWW08}).

From a mathematical point of view, SFA, LPP, genSFA and GSFA belong to the same family of optimization problems, see~\cite{ZhangEtAl-2009}, and can be solved via generalized eigenvalues. 
Conversions between these four algorithms are possible.
Two differences between LPP and GSFA are that in GSFA the vertex weights are independent of the edge weights and that GSFA is invariant to the scale of the weights, providing a normalized objective function.
 
In spite of being closely related mathematically, SFA, LPP, genSFA and GSFA originate from different backgrounds and were first motivated with different goals in mind. 
Therefore, they usually have different applications and are trained with different similarity matrices, resulting in features with different properties. For instance, LPP originates from the field of manifold learning and transitions are typically defined from input similarities (nearest neighbors).
SFA originates from unsupervised learning and transitions are typically defined by the temporal ordering of the samples.
genSFA typically combines input similarities with class information.
In GSFA, transitions typically reflect label similarities. 
For SFA and GSFA the use of input similarities might be disadvantageous, because it might compromise the invariance of the features, which is one of their central goals. 

HiGSFA is a hierarchical implementation of information-preserving GSFA (iGSFA), which has one parameter $\Delta_T$ ($\Delta$-threshold) more than GSFA. This parameter balances the number of slow and reconstructive features. When $\Delta_T < 0$ iGSFA becomes equivalent to PCA, and when $\Delta_T \ge 4.0$ iGSFA computes a linear transformation of GSFA features. Theory justifies fixing $\Delta_T$ slightly smaller than $2.0$ (Section~\ref{sec:information_loss}).

The rest of the article is organized as follows. 
In the next section we review GSFA.
In Section~\ref{sec:HSFA}, we analyze the advantages and limitations of hierarchical networks for slow feature extraction.
In Section~\ref{sec:iSFA}, we propose the iSFA (and iGSFA) algorithm. 
In Section~\ref{sec:experiments_iSFA}, we evaluate HiGSFA experimentally using the problem of age estimation from facial photographs. 
We conclude with a discussion section.

\section{Overview of Graph-based SFA (GSFA)} 
\label{sec.gSFA}
A training graph $G=(\vect{V},\vect{E})$ (illustrated in
Figure~\ref{fig.training_graph}.a) has a set $\vect{V}$ of vertices
$\vect{x}(n)$, each vertex being a sample (i.e.\ an $I$-dimensional vector),
and a set $\vect{E}$ of edges
$(\vect{x}(n), \vect{x}(n'))$, which are pairs of samples, with $1 \le
n,n' \le N$. The index $n$ (or $n'$) substitutes the time variable $t$ of SFA.
The edges are undirected and have symmetric weights $\e_{n,n'} \,=\, \e_{n',n}$, which indicate the similarity between the connected vertices;
also each vertex $\vect{x}(n)$ has an associated weight $v_n > 0$, which
can be used to reflect its frequency. 
This representation includes the standard time series of SFA as a special
case 
(Figure~\ref{fig.training_graph}.b).

\makeatletter%
\if@twocolumn%
	\newcommand{\mysizetraininggraphs}{0.70\columnwidth}
\else
	\newcommand{\mysizetraininggraphs}{0.55\columnwidth}
\fi
\makeatother

\begin{figure}[htb!]
\begin{small}
\begin{center}
\includegraphics[width=\mysizetraininggraphs]{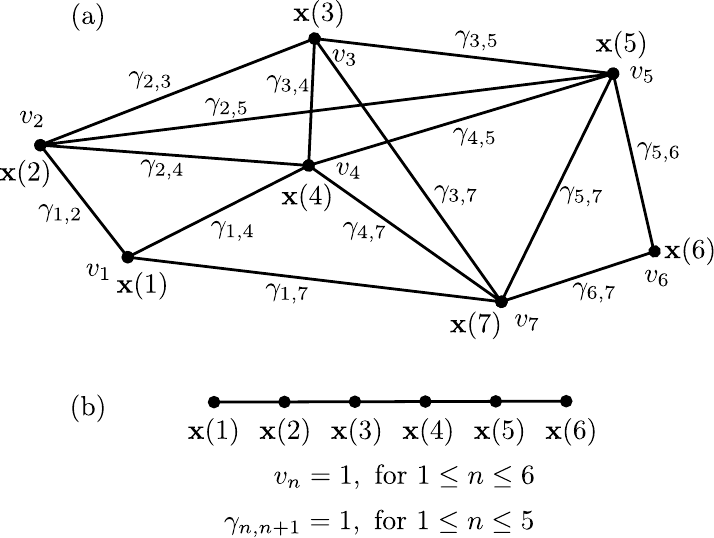}
\caption{{\tcap (a)} Example of a training graph with $N=7$ vertices. {\tcap (b)} A  linear graph suitable for GSFA that learns the same features as SFA on the time series $\vect{x}(1), \dots, \vect{x}(6)$. (Figure from \citep{EscalanteWiskott-2013b}).}
\label{fig.training_graph}
\end{center}
\end{small}
\end{figure}

The GSFA optimization problem~\citep{EscalanteWiskott-2013b} can be stated as follows. 
For $1 \le j \le J$, find features $y_j(n)=g_j(\vect{x}(n))$ with $g_j \in \mathcal{F}$, where $1 \le n \le N$ and $\mathcal{F}$ is the feature space, such that the objective function (weighted delta value)
\begin{equation} \label{eq:wobj}
\Delta_j \eqdef \frac{1}{R} \sum_{n,n'} \e_{n,n'} (y_j(n')-y_j(n))^2 \ttxt{ is minimal } 
\end{equation}
under the constraints
\begin{align} 
\label{eq:wzm}
\frac{1}{Q} \sum_{n} v_n y_j(n)  &= 0 \, ,\\
\label{eq:wuv}
\frac{1}{Q} \sum_{n} v_n (y_j(n))^2  &= 1 \, , \ttxt{ and} \\ 
\label{eq:wdec}
\frac{1}{Q} \sum_{n} v_n y_{j}(n) y_{j'}(n) &= 0 \, \ttxt{ for }  j' < j  
\end{align}
with $R \eqdef \sum_{n,n'} \e_{n,n'}$  and $Q \eqdef \sum_{n} v_n \,$. 

The constraints (\ref{eq:wzm}--\ref{eq:wdec}) are called weighted zero mean, weighted unit variance, and weighted decorrelation.
The factors $1/R$ and $1/Q$ provide invariance to the scale of the edge weights as well as to the scale of the vertex weights. 
Typically, a linear feature space is used, but the input samples are preprocessed by a nonlinear expansion function.

Depending on the training graph chosen, the properties of the features extracted by GSFA can be quite different. Training graphs for classification favor connections between samples from the same class, whereas graphs for regression favor connections between samples with similar labels. 
In Section~\ref{sec:graph_multilabels}, we show how to combine graphs for classification and regression into an efficient graph to learn age (regression),  gender (classification) and race (classification) simultaneously.

The motivation for HiSFA and HiGSFA is to correct certain disadvantages of HSFA and HGSFA while preserving their advantages. An analysis of these advantages and disadvantages is the topic of the next section.

\section{Advantages and limitations of HSFA networks}
\label{sec:HSFA}
In this section we analyze HSFA networks from the point of view of their advantages, particularly their computational complexity, and their limitations, particularly unnecessary information loss. We focus on HSFA but HGSFA is covered by extension.
Understanding the drawbacks of HSFA is crucial to justify the extensions with information preservation proposed in Section~\ref{sec:iSFA}.

\subsection{Previous work on HSFA and terminology}
The first example of HSFA appears also in the paper that first introduces the SFA algorithm \citep{Wiskott-1998a}, where it is employed as a model of the visual system for learning invariant representations.


Various contributions have continued with this biological interpretation.
In \cite{FSW07}, HSFA has been used to learn invariant features from the simulated view of a rat that moves inside a box. In conjunction with a sparseness post-processing step, the extracted features are similar to the responses of place cells in the hippocampus.

Other works have focused more on its computational efficiency compared to direct SFA.
In \cite{FranziusWilbertEtAl-2011}, HSFA has been used for object recognition from images and to estimate pose parameters of single objects moving and rotating over a homogeneous background.
In \cite{EscalanteWiskott-2013b}, an HGSFA network with 11 layers has been used to accurately find the horizontal position of faces in photographs, which is a sub-problem of face detection.



In general, HSFA networks should be directed and acyclic with arbitrary edges otherwise, but usually they are composed of multiple layers, each having a regular structure. 
Most of the HSFA networks in the literature have a similar composition. Typical differences are how the data are split into smaller chunks and the particular processing done by the SFA nodes themselves. 


The structure of the network usually follows the structure of the data. 
For instance, networks for data in 1 dimension (e.g., audio data represented as fixed length vectors) typically have a 1-dimensional structure (e.g., Figure~\ref{fig.HQSFA}), and networks for data in 2 dimensions (e.g., images) have a 2-dimensional structure. 
This idea extends to voxel data in 3 dimensions, and beyond.


For simplicity, we refer to the input data as layer 0. 
Important parameters that define the structure of a network include: (a) The {\it output dimensionality} of the nodes. (b) The {\it fan-in} of the nodes, which is the number of nodes (or data elements) in a previous layer that feed them. (c) The {\it receptive fields} of the nodes, which refer to all the elements of the input data that (directly or indirectly) provide input to a particular node. (d) The {\it stride} of a layer, which tells how far apart the inputs to adjacent nodes in a layer are. If the stride is smaller than the fan-in, then at least one node in the previous layer will feed two or more nodes in the current layer. This is called {\it receptive field overlap}.

\subsection{Advantages of HSFA}
Interestingly, hierarchical processing reduces overfitting and can be seen as a regularization method. 
Generalization is improved even when a larger number of parameters have to be learned.
This can be explained by the fact that the input to each node in the hierarchical network has the same number of samples as the original input but much smaller dimensionality, frequently leading to good generalization to unseen data in individual nodes as well as in the whole network. 
The difference in generalization between HSFA and direct SFA is more evident when polynomial expansions are involved, because the larger expanded dimensionality in direct SFA translates into stronger overfitting. 

Another interesting property of HSFA is that the nonlinearity of the nodes accumulates across the layers, so that even when using simple expansions the network as a whole may describe a highly nonlinear feature space~\citep{EscalanteWiskott-2011}. 
Depending on the data, such a complex feature space may be necessary to extract the slowest hidden parameters and to solve the supervised problem with good accuracy.

A key motivation for preferring HSFA over direct SFA is its computational efficiency. We focus here on the complexity of training rather than of feature extraction, because the former is more relevant for applications, and the latter is relatively lightweight in both cases.
Training linear SFA has a computational (time) complexity  
\begin{align}
\label{eq:comp_SFA}
T_{\ttxt{SFA}}(N, I) = \bigO(NI^2 + I^3) \, ,
\end{align}
where $N$ is the number of samples and $I$ is the input dimensionality (possibly after a nonlinear expansion).
The same complexity holds for GSFA if one uses an efficient training graph (e.g., the clustered or serial graphs, see Section~\ref{sec:graph_multilabels}), otherwise it can be as large as (for arbitrary graphs)
\begin{align}
\label{eq:comp_GSFA}
T_{\ttxt{GSFA}}(N, I) = \bigO(N^2I^2 + I^3) \, .
\end{align}

For large $I$ (i.e., high-dimensional data) direct SFA and GSFA are therefore inefficient\footnote{The problem is still feasible if $N$ is small enough so that one might apply singular value decomposition methods. However, a small number of samples $N<I$ usually results in pronounced overfitting.}. 
Their complexity can be reduced by using HSFA and HGSFA. The exact complexity depends on the structure and parameters of the hierarchical network. As we prove below, it can be linear in $I$ and $N$ for certain networks. 


\subsection{Complexity of a quadratic HSFA Network}
Although existing applications have exploited the speed of HSFA, we are not aware of any analysis of its actual complexity. 
We compute the computational complexity of a concrete quadratic HSFA (QHSFA) network for 1D data with $L$ layers, which is illustrated in Figure~\ref{fig.HQSFA}. Each node of the network performs quadratic SFA (QSFA, i.e., a quadratic expansion followed by linear SFA).
The receptive field, fan-in, and stride of the nodes in layer 1 is $k$ input values, where $k$ is a fixed parameter. In the rest of the layers the fan-in and stride of the nodes is 2 nodes. 
Assume that the total input dimensionality is $I$ and that every node reduces the dimensionality of its input from $k$ components to $k/2$ components. 

From the structure described above, it follows that the receptive fields of the nodes are non-overlapping, the network's output has $k/2$ components, the number of layers $L$ is related to $I$ and $k$:   
\begin{equation}
\label{eq:I}
I \; = \; k 2^{L-1} \, ,
\end{equation}
and the total number of nodes in the network is 
\begin{equation}
\label{eq:NumNodes}
M \; = \; 2^L-1 \, . 
\end{equation}

The input to each QSFA node has $k$ dimensions, which is increased to $k (k+3) /2$ dimensions by the quadratic expansion. Afterwards, linear SFA reduces the dimensionality to $k/2$. 
Hence, the complexity of training a single node is 
\begin{align}
T_{\ttxt{QSFA}}(N,k) \; &\stackrel{\mathclap{(\ref{eq:comp_SFA})}}{=} \; \bigO(N (k (k+3) / 2) ^2 + (k (k+3) / 2)^3) \\
\label{eq:comp_qsfa}
&= \; \bigO(N k^4 + k^6) \, .
\end{align}
The number of nodes is $M \stackrel{(\ref{eq:I},\ref{eq:NumNodes})}{=} \bigO(I/k)$. Therefore, the complexity of training the whole network is 
\begin{align}
\label{eq:comp_QHSFA}
T_{\ttxt{QHSFA}}(N,I,k) \stackrel{(\ref{eq:comp_qsfa})}{=} \bigO((N k^4 + k^6)I/k) = \bigO(I N k^3 + I k^5) \, .
\end{align}

This means the complexity of the QHSFA network above is linear w.r.t.\ the input dimension $I$, whereas the complexity of direct QSFA is
\begin{align}
T_{\ttxt{QSFA}}(N,I) \stackrel{(\ref{eq:comp_qsfa})}{=} \bigO(N I^4 + I^6) \, ,
\end{align}
which is linear w.r.t.\ $I^6$. 
Thus, the QHSFA network has a large computational advantage over direct QSFA.


Since each layer in the QHSFA network is quadratic, in general the output features of layer $l$ can be written as polynomials of degree $2^l$. In particular, the output features of the network are polynomials of degree $2^L$.
However, the actual feature space does not span all the polynomials of this degree but only a subset of them due to the restricted connectivity of the network.
In contrast, direct QSFA only contains quadratic polynomials (although all of them).

One could train direct SFA on data expanded by a polynomial expansion of degree $2^L$. The expanded dimensionality would be $\sum_{d=0}^{2^L} {{d+I-1}\choose{I-1}} = \bigO(I^{2^L})$, resulting in a complexity of $\bigO(NI^{2^{L+1}} + I^{3 \cdot 2^L})$, being even less feasible than direct QSFA.


\makeatletter%
\if@twocolumn%
	\newcommand{\mysizeqsfa}{0.68\columnwidth}
\else
	\newcommand{\mysizeqsfa}{0.55\columnwidth}
\fi
\makeatother

\begin{figure}[htb!]
\begin{small}
\begin{center}
\includegraphics[width=\mysizeqsfa]{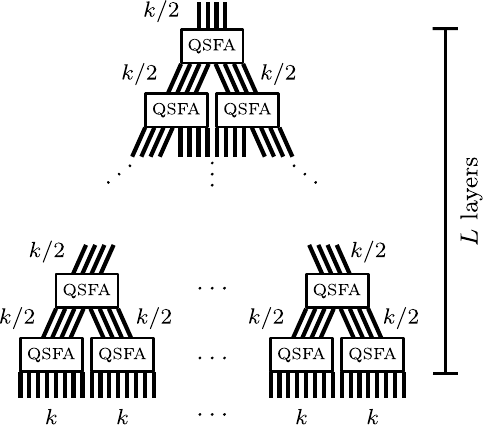}
\caption{
Example of an 1D QHSFA network with binary fan-ins and no overlap. Each node performs quadratic SFA reducing the dimensionality from $k$ to $k/2$ components. A small fan-in results in a network with a large number $L$ of layers, which may be useful to build deep networks.
}
\label{fig.HQSFA}
\end{center}
\end{small}
\end{figure}

The space (memory) complexity of linear SFA is 
\begin{align}
S_{\ttxt{SFA}}(N,I) = \bigO(I^2+ NI) \, 
\end{align}
(where the term $NI$ is due to the input data). One can reduce this by using HSFA and training each node separately, one at a time. 
For instance, the space complexity of direct QSFA is 
\begin{align}
S_{\ttxt{QSFA}}(N,I) = \bigO(I^4+ NI) \, ,
\end{align}
whereas the space complexity of the QHSFA network is only
\begin{align}
S_{\ttxt{QHSFA}}(N,I,k) = \bigO(k^4+NI) \, .
\end{align}

It is possible to design more sophisticated networks than the QHSFA one and preserve a similar computational complexity.
For example, the network proposed in Section~\ref{sec:experiments_iSFA} has overlapping receptive fields, larger fan-ins, a 2D structure, and a complexity also linear in $I$ and $N$.

\subsection{Limitations of HSFA networks}
\label{sec:limitations_HSFA}
In this section, we analyze the limitations of HSFA networks, or in general, any network in which the nodes have only one criterion for DR, namely, slowness maximization.
Otherwise the nodes are unrestricted; they might be linear or nonlinear, include additive noise, clipping, various passes of SFA, etc. 

In spite of the remarkable advantages of HSFA, we show that relying only on slowness to determine which aspects of the data that are preserved results in three disadvantages: unnecessary information loss, poor input reconstruction and feature garbling. 

\paragraph{Unnecessary information loss}
\label{sec:information_loss}
This disadvantage occurs when the nodes discard dimensions of the data that are not significantly slow locally (i.e., at the node level), but which would have been useful for slowness optimization at higher layers of the network if they had been preserved. 

\begin{sloppypar}
We show through a toy experiment that dimensions important for global slowness are not necessarily slow locally. 
Consider four zero-mean, unit-variance signals: $s_1(t)$, $s_2(t)$, $s_3(t)$ and $n(t)$ that can only take binary values in $\{-1, +1 \}$ ($n$ stands for noise here), where $\Delta_{s_1} < \Delta_{s_2} < \Delta_{s_3} < \Delta_{n} = 2.0$, and all features are independent of the rest\footnote{One can show that the expected $\Delta$ value of a random unit-variance i.i.d.\ noise feature is 2.0~\cite{EscalanteWiskott-2015a}. The same holds for GSFA if the graph is consistent and has no self-loops.}.
Assume the 4-dimensional input is $(x_1,x_2,x_3,x_4) \eqdef (s_2, s_1 n, s_3, n)$ and the number of samples is large enough. Thus, the slowest possible feature described by the data is $x_2 x_4 = (s_1 n) n = s_1$ (or equivalently $-x_2 x_4$). 
\end{sloppypar}

QSFA would be able to extract the slowest feature from these data, but let us assume that a 2-layer QHSFA network with 3 nodes is used, where the output feature is:
$\ttxt{QSFA}\big( \ttxt{QSFA}(s_2, s_1 n), \ttxt{QSFA}(s_3, n) \big)$.
Each QSFA node reduces the number of dimensions from 2 to 1. 
Since $\Delta_{s_2} < \Delta_{s_1 n} = 2.0$, the first bottom node computes $\ttxt{QSFA}(s_2, s_1 n)=s_2$, and since $\Delta_{s_3} < \Delta_{n} = 2.0$, the second bottom node computes $\ttxt{QSFA}(s_3, n)=s_3$. The top node extracts $\ttxt{QSFA}(s_2, s_3) = s_2$. Therefore, the network misses the slowest feature, $s_1$, even though it belongs to the feature space spanned by the network.

The problem can be expressed in information theoretic terms: 
\begin{align}
I(s_1 n, s_1) &= 0 \, \ttxt{, and} \\
I(n, s_1) &= 0 \, \ttxt{, but} \\
I((s_1 n, n), s_1) &= H(s_1) > 0 \, ,
\end{align}
where $I$ denotes mutual information, and $H$ is entropy. 
Thus, it is impossible to locally rule out that a feature contains information that might yield a slow feature, unless one (globally) observes the whole data available to the network.

\makeatletter%
\if@twocolumn%
	\newcommand{\mysizedeltas}{0.90\columnwidth}
\else
	\newcommand{\mysizedeltas}{0.70\columnwidth}
\fi
\makeatother

\begin{figure}[htb!]
\begin{small}
\begin{center}
\includegraphics[width=\mysizedeltas]{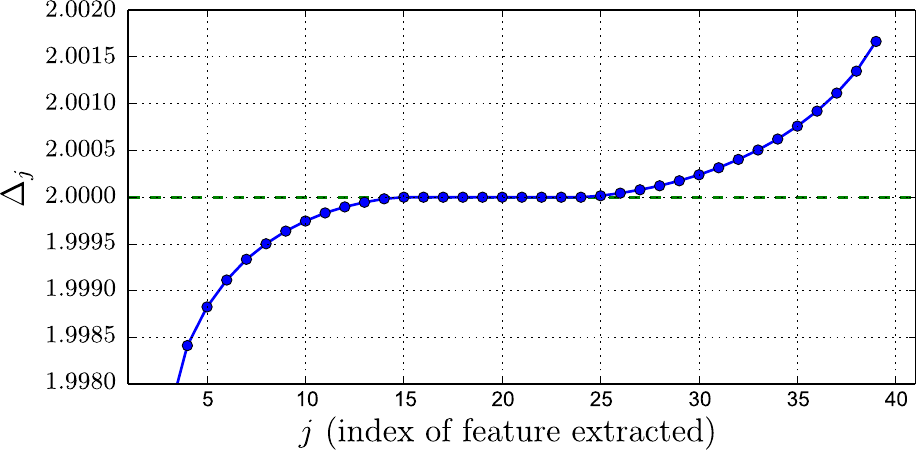}
\caption{
$\Delta$ values of the first 40 slow features averaged over all the nodes of the first layer of an HGSFA network trained for age estimation ($\Delta_1=1.859$, $\Delta_2=1.981$, and $\Delta_3=1.995$, not shown). 
The training graph employed is a serial graph (see Section \ref{sec:graph_multilabels}) with 32 groups.
Most $\Delta$ values are close to $2.0$, indicating that at this early stage, where the nodes have small (6$\times$6-pixel) receptive fields, the slowest features found are not significantly slow.
}
\label{fig.delta_values}
\end{center}
\end{small}
\end{figure}

Unnecessary information loss can also affect applications in practice.
For example, Figure~\ref{fig.delta_values} shows the $\Delta$ values of the slowest features extracted by the first layer of an HGSFA network trained for age estimation on human face images. 
Most $\Delta$ values are approximately 2.0, and only a few of them are less than 2.0.
%

A feature with $\Delta = 2.0$ can be a transformation of the input data, a transformation of inherent noise, or a mixture of both.
In fact, if two or more feasible features have the same $\Delta$ value, SFA outputs an arbitrary rotation of them, even though one might prefer features that are transformations of the input only rather than noise.
Due to the DR only a few features with $\Delta = 2.0$ may be preserved. The discarded features might still contain useful information, giving place to information loss.

\paragraph{Poor input reconstruction}
The goal of input reconstruction is to generate an input from a given slow feature representation.
This interesting task has been studied in~\cite{Wilbert-2012}, and may help to determine which features the network is sensitive to, or to find inputs with certain properties. 

In the field of image processing, reconstruction might be relevant for image morphing and interpolation. 
Here, we consider morphing as the task of finding how the input must be modified to reflect modifications introduced to the output features.
For example, assume SFA was trained to extract age from facial images.
Morphing would allow us to visualize how a particular person would look a few years older or younger.

Experiments have shown that input reconstruction from the top-level features extracted by HSFA is a challenging task \cite{Wilbert-2012}. 
We have also conducted experiments with various nonlinear methods for input reconstruction, including local and global methods, confirming its difficulty.

We show here that the cause of poor input reconstruction may not be the weakness of the reconstruction algorithms employed but insufficient reconstructive information in the slow features. 
The extracted features ideally depend only on the hidden slow parameters and are invariant to any other factor.
In the age estimation example, if the extracted features are close to the features predicted from theory, they would be strongly related to the age and harmonic functions of it. Thus, they would be mostly invariant to other factors, such as the identity of the person, his facial expression, the background, etc. Therefore, in theory only the age would be available for reconstruction. 

In practice, the features usually contain residual information about the input data. However, one cannot rely on this information because it may be partial, making reconstructions not unique, and highly nonlinear, making it difficult to untangle it.
Even the features extracted with linear SFA typically result in inaccurate reconstructions.
Using many layers of SFA may make the problem more serious in HSFA.

One exception where reconstruction is possible is when SFA is trained with artificial data generated only from slowly changing parameters. In this case, the slow output features should encode the generative parameters, allowing input reconstruction if the reconstruction method is powerful enough.

\paragraph{Feature garbling}
Even if a node does not incur in information loss, the features might still not be represented in a useful way for the extraction of slow features in the next nodes. 
Feature garbling occurs when the features extracted by SFA are more complex (e.g., highly nonlinear) than necessary without being slower. 
This complicates the extraction of slow features in higher layers, because the feature space available to the nodes might not suffice to transform the representation into slow features.

We exemplify feature garbling with another toy experiment. Consider two feature vectors $\vect{x}(t)$ and $\vect{x}'(t) = E_k(\vect{x}'(t))$, where $E_k$ is a (bijective) encryption function for some unknown key $k$. 
Assume that $\Delta(x_1(t)) = \cdots = \Delta(x_I(t)) = 2.0$, $\Delta(x'_1(t)) = \cdots = \Delta(x'_I(t)) = 2.0$, and that $\vect{x}(t)$ and $\vect{x}'(t)$ are valid output signals that belong to the feature space.
Since both signals have the same $\Delta$-values, SFA might output any of them. 
The signal $\vect{x}(t)$ might be more useful for slowness maximization in the next layers because its features are simpler and more directly connected to the inputs, where as $\vect{x}'(t)$ might be useless for the next layers, unless the decryption function belongs to the feature space.

Feature garbling results in information loss. An informative but garbled feature is likely to be discarded later in the hierarchy.
One might attempt to reduce feature garbling by using a mild nonlinear expansion in SFA. However, this might compromise the slowness of the extracted features.
Similarly, one might try to preserve a large number of features to reduce information loss. However, this might be impractical because it would increase the computational cost and contradicts the goal of dimensionality reduction.

The problems of poor input reconstruction and unnecessary information loss are also connected. This is evident if one distinguishes between two types of information: (a) information about the full input data and (b) information about the global slow parameters. 
Losing (a) results in poor input reconstruction, whereas losing (b) results in unnecessary information loss. Of course, (a) contains (b). Therefore, both problems originate from losing different but related types of information. 

Feature garbling is a theoretical open issue that still needs to be formalized and whose relevance in practice is unknown. Therefore, we do not attempt to counteract it with the proposed extensions.

\section{Information-preserving SFA (iSFA)}
\label{sec:iSFA}
In this section, we propose information-preserving SFA (iSFA) to counteract the problems of unnecessary information loss and poor input reconstruction.
We write iSFA with lowercase `i' to distinguish it from independent SFA (ISFA)~\cite{BlasZitoWisk2007}. 
The extension can also be applied to GSFA and is then called information-preserving GSFA (iGSFA). For simplicity, we first focus on iSFA. 

iSFA combines two learning principles: the slowness principle and information preservation without compromising the former in any way. 
Information preservation requires the maximization of the mutual information between the output features and the input data. 
However, 
for finite, discrete and typically unique data samples 
it is difficult to measure and maximize mutual information unless one assumes some probability model. Therefore, we interpret information preservation more practically and minimize a reconstruction error.
A closely related concept is the explained variance, but we avoid this term because it is typically restricted to linear transformations.
When iSFA is used in hierarchical networks, the concept of information preservation might also be called information propagation.

In the rest of the section, first we present a high-level description of the algorithm, then, we describe its construction in detail, show how to approximate an inverse transformation, and discuss its main properties.

\subsection{Algorithm overview}
\label{sec:alg_overview}
The goal of iSFA is to improve the feature extraction of HSFA networks by improving feature extraction at the node level, which is achieved by replacing the SFA nodes with iSFA nodes.
Therefore, the structure of the HSFA network can be preserved.

An essential property of iSFA is that the feature vectors are composed of two parts: (1) a slow part derived from SFA features, and (2) a reconstructive part derived from principal components (PCs). 

Roughly speaking, the slow part captures the slow aspects of the data and is basically composed of standard SFA features, except for an additional linear mixing step explained in Sections~\ref{sec:alg_sfa_pc} and \ref{sec:mixing_feats}. 
The reconstructive part ignores the slowness criteria and instead focuses on describing the input linearly as closely as possible (disregarding the part already described by the slow part). 
In Section~\ref{sec:experiments_iSFA} we show that, although the reconstructive features are not particularly slow, they contribute to global slowness maximization.


The proposed algorithm takes care of the following important considerations.
(a) Given the output dimension $D$, it decides how many features the slow and reconstructive part should contain.
(b) It minimizes the redundancy between the slow and the reconstructive part, allowing the output features to be more compact and have higher information content.
(c) It corrects the amplitudes of the slow features (SFA features usually have unit variance) to make them compatible with the PCs (PCA is a rotation and projection, preserving thus the amplitude of the original data).


\subsection{Algorithm description (training phase)}
\label{sec:alg_sfa_pc}
In this section, we present iSFA in detail, or more precisely, its training phase.
Figure~\ref{fig.iSFA} shows the components involved in the algorithm, and Algorithm~\ref{alg:iSFA.train} gives a concise description.


\makeatletter%
\if@twocolumn%
	\newcommand{\mysizesfapc}{0.80\textwidth}
\else
	\newcommand{\mysizesfapc}{0.93\textwidth}
\fi
\makeatother

\begin{figure*}[htb!]
\begin{small}
\begin{center}
\includegraphics[width=\mysizesfapc]{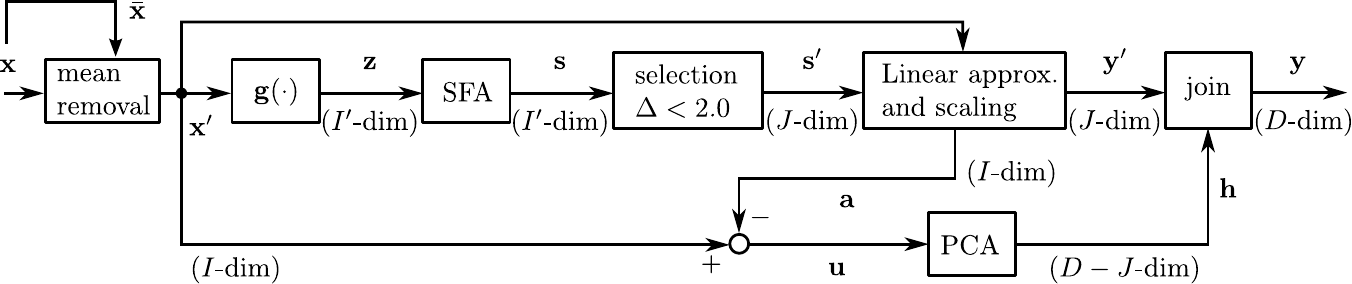}
\caption{
Block diagram of the iSFA node showing the components used for training, feature extraction, and linear input reconstruction. The blocks and signals are explained in the text.
}
\label{fig.iSFA}
\end{center}
\end{small}
\end{figure*}


Let $\vect{X} \eqdef (\vect{x}_1, \dots, \vect{x}_N)$ be the $I$-dimensional training data, $D$ the output dimensionality, $g(\cdot)$ the expansion function, and $\Delta_T=2.0$ (a $\Delta$-threshold, in practice slightly smaller than 2.0). 
First, the average sample $\bar{\vect{x}} \eqdef \frac{1}{N}\sum_n \vect{x}_n$ is removed from the $N$ training samples resulting in the centered data $\vect{X}' \eqdef \{ \vect{x}'_n \}$, with $\vect{x}'_n \eqdef \vect{x}_n - \bar{\vect{x}}$, where $1 \le n \le N$. 
Then, $\vect{X}'$ is expanded via $\vect{g}(\cdot)$, resulting in vectors $\vect{z}_n=\vect{g}(\vect{x'}_n)$ of dimensionality $I'$.
Afterwards, a new instance of linear SFA is created and trained with the expanded  data $\vect{Z} \eqdef \{ \vect{z}_n \}$. The slow features extracted from the expanded data are denoted $\vect{s}_n$.
The first $J$ components of $\vect{s}_n$ with $\Delta < \Delta_T$ and $J < J' \eqdef \min(I', D)$, are denoted $\vect{s}'_n$. The remaining slow features are discarded.


The next steps correct the amplitude of the $J$ slow features in $\vect{S}'\eqdef \{ \vect{s}'_n \}$, which have unit variance. 
The centered data $\vect{X}'$ is approximated linearly from $\vect{S}'$ by computing a matrix $\vect{M}$ and a vector $\vect{b}$, as follows. 
\begin{align}
\label{eq:A_inMSp}
\vect{A} \eqdef \vect{M}\vect{S'}+\vect{b}\vect{1}^T \approx \vect{X'} \, ,
\end{align}
where $\vect{A}$ is the contribution of the slow part to the approximation of the centered data (i.e., $\vect{x}'_n \approx \vect{a}_n \eqdef \vect{M}\vect{s}'_n+\vect{b}$) and $\vect{1}$ is a vector of 1s of length $N$.
Since $\vect{X}'$ and $\vect{S}'$ are centered, $\vect{b}$ could be discarded because $\vect{b}=\vect{0}$. However, when GSFA is used the slow features have {\it weighted} zero mean, and $\vect{b}$ might improve the approximation of $\vect{X}'$.
The QR decomposition of $\vect{M}$ is computed as 
\begin{align}
\label{eq:M_inQR}
\vect{M} = \vect{Q} \vect{R} \, ,
\end{align}
where $\vect{Q}$ is orthonormal and $\vect{R}$ is upper triangular. Then, the slow feature part is computed as 
\begin{align}
\label{eq:yp_inRsp}
\vect{y}'_n= \vect{R} \vect{s}'_n \, .
\end{align}
Section~\ref{sec:mixing_feats} justifies this mixing and scaling of the slow features $\vect{s}'_n$.

To obtain the reconstructive part, 
the residual data $\vect{u}_n \eqdef \vect{x}'_n - \vect{a}_n$ are computed, i.e., the data that remains after the data linearly reconstructed from $\vect{y}'_n$ (or $\vect{s}'_n$) is removed from the centered data. 
Afterwards, a new PCA instance is trained with $\{ \vect{u}_n \}$. Then, the reconstructive part $\vect{h}_n$ is defined as the $D-J$ first principal components of $\vect{u}_n$. 

Afterwards, the concatenation of the slow part $\vect{y'_n}$ ($J$ features) and reconstructive part $\vect{h_n}$ ($D-J$ features) results in the $D$-dimensional output features $\vect{y_n} \eqdef \vect{y'_n} | \vect{h_n}$, where $|$ is vector concatenation.

Finally, the algorithm returns $\vect{Y}=(\vect{y_1}, \dots, \vect{y_N})$, $\vect{\bar{x}}$, $J$, $\vect{Q}$, $\vect{R}$, $\vect{b}$, and the trained $\tsys{SFA}$ and $\tsys{PCA}$ instances.
The output features $\vect{Y}$ are usually computed only during feature extraction (see Algorithm~\ref{alg:iSFA.execute}). Still, we keep them here to simplify the understanding of the signals involved. 

\algnewcommand{\LineComment}[1]{\( // \) #1}
\newcommand\RComment[1]{\LineComment{\minibox[l]{#1}}}

\begin{algorithm}
\caption{Training phase of iSFA}\label{alg:iSFA.train}
\begin{algorithmic}[1]
\Require $D>0$: output dimension
\Procedure{train}{$\vect{X}$} \Comment{$\vect{X}=(\vect{x_1}, \dots, \vect{x_N})$: training samples } 
   \State $\forall n: \vect{x'_n} \gets \vect{x_n}-\vect{\bar{x}}$ \Comment{$\vect{\bar{x}}$: average sample}
   \State $\forall n: \vect{z_n} \gets \vect{g}(\vect{x'_n})$ 
   \State $\vect{W}_\ttxt{SFA}, \bar{\vect{z}} \gets \tsys{SFA.train}(\vect{Z}, \tsys{output\_dim}=J')$ \Comment{$\vect{Z}=(\vect{z_1}, \dots, \vect{z_N})$: expanded samples}
   \State $\forall n: \vect{s_n} \gets \vect{W}_\ttxt{SFA} (\vect{z_n}-\bar{\vect{z}})$ 
   \State $\forall n: \vect{s'_n} \gets (\vect{s_{n1}}, \dots, \vect{s_{nJ}})^T$  \Comment{Preserve the first $J$ features with $\Delta < \Delta_T$}

   \State $\forall n: \vect{a_n} \gets \vect{M}\vect{s'_n} + \vect{b}$ \Comment{For $\vect{M},\vect{b}$, such that $\vect{M}\vect{S'} + \vect{b} \approx \vect{X'}$}
   \State $\forall n: \vect{y'_n} \gets \vect{R}\vect{s'_n}$ 
   \Comment{For $\vect{Q}\vect{R} = \vect{M}$, the QR decomposition of $\vect{M}$}

   \State $\forall n:  \vect{u_n} \gets \vect{x'_n} - \vect{a_n}$ 
   

   \State $\vect{W}_\ttxt{PCA} \gets \tsys{PCA.train}(\vect{U}, \tsys{output\_dim}=D-J)$ 
   \State $\forall n: \vect{h_n} \gets \vect{W}_\ttxt{PCA} \vect{u_n}$ \Comment{Only $D-J$ PCs are preserved}
   \State $\forall n: \vect{y_n} = \vect{y'_n} | \vect{h_n}$ \Comment{`$|$' denotes vector concatenation}

   \State \textbf{return} $\vect{Y}=(\vect{y_1}, \dots, \vect{y_N})$, $\vect{\bar{x}}$, $\vect{W}_\ttxt{SFA}$, $\bar{\vect{z}}$, $\vect{W}_\ttxt{PCA}$, $J$, $\vect{Q}$, $\vect{R}$, $\vect{b}$
\EndProcedure
\end{algorithmic}
\end{algorithm} 

\subsection{Feature extraction}
The algorithm for feature extraction is similar to the training algorithm, except that the parameters $\vect{\bar{x}}$, $\vect{W}_\ttxt{SFA}$, $\bar{\vect{z}}$, $\vect{W}_\ttxt{PCA}$, $J$, $\vect{Q}$, $\vect{R}$, $\vect{b}$ have already been learned from the training data. 
Algorithm \ref{alg:iSFA.execute} shows how a single input sample is processed, however, it can be easily and efficiently adapted to process multiple input samples by taking advantage of matrix operations.

One interesting property of iSFA is that the features are nonlinear w.r.t.\ the input data, both the slow and the reconstructive part.
The slow part is nonlinear due to the expansion function. 
The residual data are nonlinear because it is computed using the (nonlinear) slow part and the centered data. 
The reconstructive part is thus only linear w.r.t the residual data but nonlinear w.r.t.\ the input data. 

\begin{algorithm}
\caption{Feature extraction with iSFA}\label{alg:iSFA.execute}
\begin{algorithmic}[1]
\Require $D$, $\vect{\bar{x}}$, $\vect{W}_\ttxt{SFA}$, $\bar{\vect{z}}$, $\vect{W}_\ttxt{PCA}$, $J$, $\vect{Q}$, $\vect{R}$, $\vect{b}$
\Procedure{extract}{$\vect{x}$} \Comment{$\vect{x}$: a new sample}
%
   \State $\vect{x'} \gets \vect{x}-\vect{\bar{x}}$ \Comment{$\vect{\bar{x}}$: mean of the training data}
   \State $\vect{z} \gets \vect{g}(\vect{x'})$ 
   \State $\vect{s} \gets \vect{W}_\ttxt{SFA} (\vect{z_n}-\bar{\vect{z}})$ \Comment{Extract the first $J$ slow features} 
   \State $	\vect{s'} = (s_1, s_2, \dots, s_J)$
   \State $\vect{y'} \gets \vect{R}\vect{s'}$ 
   \State $\vect{a} \gets \vect{Q}\vect{y'}+\vect{b}$ 
   \State $\vect{u} \gets \vect{x'} - \vect{a}$ 
   \State $\vect{h} \gets \vect{W}_\ttxt{PCA} \vect{u}$ 
   \State $\vect{y} = \vect{y'} | \vect{h}$ 

   \State \textbf{return} $\vect{y}$
\EndProcedure
\end{algorithmic}
\end{algorithm}

\subsection{Mixing and scaling of slow features}
\label{sec:mixing_feats}
In the iSFA algorithm, the $J$-dimensional slow features $\vect{s}'_n$ are transformed into the scaled $\vect{y}'$ features.
This transformation is necessary to make the amplitude of the slow features compatible with the amplitude of the PCA features, so that PCA processing of both sets of features together is possible in the next layers.
The \emph{QR scaling}, used by Algorithm~\ref{alg:iSFA.train}, as well as a \emph{sensitivity-based scaling}, are explained below.


Both methods ensure that the amplitude of the slow features is approximately equal to the reduction in the reconstruction error that they allow. In practice, a lower bound on the scales (not in the pseudo-code) ensures that all features have amplitudes $> 0$ even if they do not contribute to reconstruction.

A scaling method should ideally offer two key properties of PCA. (1) If one adds unit-variance noise to one of the principal components, the variance of the reconstruction error also increases by one unit. (2) If one adds independent noise to two or more principal components simultaneously, the variance of the reconstruction error increases additively. 

Since $\vect{Q}$ is orthogonal, $\vect{y'}(t) \stackrel{(\ref{eq:A_inMSp},\ref{eq:M_inQR},\ref{eq:yp_inRsp})}{=} \vect{Q}^T (\vect{a}(t) - \vect{b})$ and $\vect{y'}$ is a rotation of the reconstructed data $\vect{a}$ (after centering). Thus, $\vect{y'}$ fulfills the two key properties of reconstruction of PCA w.r.t.\ $\vect{a}$. 

One small drawback is that $(\ref{eq:yp_inRsp})$ mixes the slow features.
Polynomial expansion functions combined with SFA are invariant to invertible linear transformations (e.g.\ $\ttxt{SFA}(\allowbreak \tsys{QExp}(\vect{U}\vect{x})) \equiv \ttxt{SFA}(\ttxt{QExp}(\vect{x}))$, where $\vect{U}$ is any invertible matrix and $\tsys{QExp}$ is the quadratic expansion). 
Thus, polynomial SFA can extract the same features from $\vect{s}'$ or $\vect{y}'$. 
However, other expansions do not have this property. One example of them is the function
\begin{align}
\label{eq:08expo}
\tsys{0.8Exp}(x_1, x_2, \dots, x_I) &\eqdef \notag \\
(x_1, x_2, \dots, x_I, |x_1&|^{0.8}, |x_2|^{0.8}, \dots, |x_I|^{0.8}) \, ,
\end{align}
which when combined with SFA is invariant to scalings of the input data but not to their mixing, i.e., $\ttxt{SFA}(\tsys{0.8Exp}(\vect{\Lambda}\vect{x})) \equiv \ttxt{SFA}(\tsys{0.8Exp}(\vect{x}))$, where $\vect{\Lambda}$ is a diagonal matrix with diagonal elements $\lambda_i \neq 0$, but $\ttxt{SFA}(\tsys{0.8Exp}(\vect{U} \vect{x})) \not\equiv \ttxt{SFA}(\tsys{0.8Exp}(\vect{x}))$ in general. 
The $\tsys{0.8Exp}$ expansion has been motivated by a model where the slow features are noisy harmonics of increasing frequency of a hidden parameter and it should be applied to the slow features directly. 
Thus, the mixture of the slow features would break the assumed model, which might compromise slowness extraction in the next layers.


Technically, such mixing could be reverted in the next layer (e.g., by an additional application of linear SFA before the expansion), but this would add unnecessary complexity. 
For this reason, besides the QR scaling, we propose a second scaling method. The sensitivity based scaling scales the slow features without mixing them, as follows. 
\begin{align}
\label{eq:yp_inLsp}
\vect{y'}=\vect{\Lambda} \vect{s'} \, ,
\end{align}
where $\vect{\Lambda}$ is a diagonal matrix with diagonal elements $\lambda_j \eqdef || \vect{M}_j ||_2$ (the $L_2$-norm of the j-th column vector of $\vect{M}$). 
Therefore, 
\begin{align}
\vect{a}(t) \stackrel{(\ref{eq:yp_inLsp})}{=} \vect{M} \vect{\Lambda}^{-1} \vect{y}'(t) + \vect{b} \, .
\end{align}

Clearly, the transformation (\ref{eq:yp_inLsp}) does not mix the slow features, it only scales them. 
From the two key reconstruction properties of PCA mentioned above 
(adding noise of certain variance to either one or many features increases the variance of the reconstruction error by the same amount), the first one (noise on a single feature) is fulfilled, because the columns of $\vect{M} \vect{\Lambda}^{-1}$ have unit norm. 
The second one is not fulfilled, because $\vect{M}\vect{\Lambda}^{-1}$ is in general not orthogonal. 

On a first sight, it seems like multiple-view learning might be an alternative for the scaling methods used here. However, it is actually solves a different problem. Moreover, it is not suitable for this task because it would mix the slow and reconstructive parts, and might be expensive computationally (e.g., \cite{XiaEtAl-2010}) with cubic complexity in $N$.



\subsection{Input reconstruction for iSFA}
\label{sec:inverse_iSFA}
iSFA allows a linear approximation of the input  (linear input reconstruction) even though all the features computed are nonlinear. 
In contrast, standard SFA does not have a standard input reconstruction method, although various gradient-descent and vector-quantization methods have been tried (e.g., \citep{Wilbert-2012}) with limited success.

In the reconstruction algorithm, $\vect{a}$ (the contribution of the slow part to the centered data) is approximated as $\vect{\tilde{a}} = \vect{Q}\vect{y'}+\vect{b}$, where tilde denotes an approximation.
Then, $\vect{u}$ (the residual vector) is approximated as $\vect{\tilde{u}} = \vect{W}_{\ttxt{PCA}}^T \vect{h}$.
The reconstructed sample is then $\vect{\tilde{x}} = \vect{\tilde{a}} + \vect{\tilde{u}} + \vect{\bar{x}}$.
See Algorithm~\ref{alg:iSFA.inverse} for details.

\begin{algorithm}
\caption{Linear input reconstruction for iSFA}\label{alg:iSFA.inverse}
\begin{algorithmic}[1]
\Require $D$, $\vect{\bar{x}}$, $\vect{W}_\ttxt{PCA}$, $J$, $\vect{Q}$, $\vect{R}$, $\vect{b}$
\Procedure{linear-reconstruction}{$\vect{y}$} 
  \State $\vect{y'} \gets (y_1, \dots, y_J)$ \Comment{Slow part} 
  \State $\vect{h} \gets (y_{J+1}, \dots, y_D)$ \Comment{Reconstructive part} 
  \State $\vect{\tilde{x}} \gets (\vect{Q}\vect{y'}+\vect{b}) + \vect{W}_{\ttxt{PCA}}^T \vect{h} + \vect{\bar{x}}$ \Comment{$\vect{\tilde{a}} + \vect{\tilde{u}} + \vect{\bar{x}}$}
  \State \textbf{return} $\vect{\tilde{x}}$
\EndProcedure
\end{algorithmic}
\end{algorithm}

The linear reconstruction algorithm has interesting properties. Firstly, it is simpler than the feature extraction algorithm, because the nonlinear expansion and $\vect{W}_\ttxt{SFA}$ are not needed.
Secondly, it has a low computational complexity, because it consists of only two matrix-vector multiplications and a couple of vector additions, all of them $I$, $J$, $D-J$ or $D$-dimensional. 

Linear reconstruction is simple and effective for iSFA. We now describe a nonlinear reconstruction algorithm. Assume $\vect{y}$ is the iSFA feature representation of a sample $\vect{x}$, which we denote as $\vect{y} = \ttxt{iSFA}(\vect{x})$. 
Since $\vect{x}$ is unknown, the reconstruction error cannot be computed directly.
However, one can indirectly measure the accuracy of a particular reconstruction $\vect{\tilde{x}}$ by means of the feature error, which is defined here as $e_\ttxt{feat} \eqdef || \vect{y} - \ttxt{iSFA}(\vect{\tilde{x}}) ||$. 
%
%
This feature error can be minimized for $\vect{\tilde{x}}$ using the function $\ttxt{iSFA}(\cdot)$ as a black box and gradient descent or other generic nonlinear minimization algorithms. Frequently, such algorithms require a first approximation, which can be very conveniently provided by the linear reconstruction algorithm.

Although nonlinear reconstruction methods might result in higher reconstruction accuracy, they are typically more expensive computationally. Moreover, in the discussion we explain why minimizing $e_\ttxt{feat}$ does not necessarily improve the reconstruction error unless other aspects are considered. 


\subsection{Some remarks on iSFA}
Clearly, the computational complexity of iSFA is at least that of SFA, because iSFA consists of SFA and a few additional computations.
However, none of these additional computations is done on the expanded $I'$-dimensional data but at most on $I$ or $D$-dimensional data (e.g., PCA is applied to $I$-dimensional data, and the QR decomposition is applied to a $I \times I$-matrix resulting in $\bigO(NI^2+I^3)$ and $\bigO(I^3)$ complexity, respectively).
Therefore, iSFA is slightly slower than SFA but it has the same complexity order.
Practical experiments (Section \ref{sec:experiments_iSFA}) confirm this.  
 
The presentation above focuses on iSFA. To obtain information-preserving GSFA (iGSFA) one only needs to substitute SFA by GSFA inside the iSFA algorithm and provide GSFA with the corresponding training graph during the training phase.
Notice that the GSFA features have weighted zero-mean instead of the simple (unweighted) zero-mean enforced by SFA. 
This small difference has already been compensated by the vector $\vect{b}$.

We have implemented the iGSFA node (including iSFA) in python. In the next few months, we plan to make the node public by integrating it to the MDP toolkit~\cite{ZitoEtAl-MDP-2009}.

\section{Experimental evaluation}
\label{sec:experiments_iSFA}
In this section, we evaluate HiGSFA using the problem of age estimation from human face photographs. HiGSFA is employed instead of HiSFA, because we use the labels to boost estimation accuracy.
However, due to their close connection, many aspects of the evaluation also extend to HiSFA.

First, age estimation and previous work are introduced. Then, the pre-processing of the images is described. Afterwards, a new training graph for learning age, race and gender is proposed. Then, an HiGSFA network is described and evaluated according to three criteria: feature slowness (compared with HGSFA), age estimation error (compared with state-of-the-art algorithms), and linear reconstruction error (compared with PCA). 

\subsection{Age estimation and previous work on this problem}
The study of age estimation from photographs is relatively recent and has useful applications for human-computer interaction, group-targeted advertisement, demographics, face recognition, control of age-related policies, and security. 
However, age estimation is a challenging task probably because different persons experience facial aging differently depending on factors such as their gender, race, life style, nutrition, health, exposure to the weather, use of creams or cosmetics, operations, accidents, and even psychological traits. 

Two types of age have been distinguished. The real (ground-truth) age, which is the chronological age of the person, and the apparent age, which is the age conveyed solely by the information present in the image. Clearly, an algorithm cannot determine the real age exactly, at most it might determine the apparent age. 

\label{sec:literature_age_estimation}
Different methods for age estimation have been proposed. 
For a more comprehensive literature review we refer to \cite{RamanathanChellappaBiswas-2009} and \cite{FuGuoHuang-2010}.
In \cite{GengZhouSmith-2007}, aging pattern subspace (AGES) has been proposed, which is based on temporal sequences of images of individual persons, the so-called aging patterns. 
The images are represented through an appearance model that combines geometric and texture information.
In their system, a subspace is constructed for each aging pattern. 
Given a new image, the subspace providing the best possible reconstruction is found. Then, the position of the image within the aging pattern is determined.

Guo et al.\ \cite{GuoMuFuHuang-2009} have proposed the use of bio-inspired features (BIF). Their architecture consists of two-layers, in which the units of the first layer compute Gabor functions inspired by simple cells, whereas the units of the second layer compute a standard-deviation operation inspired by complex cells. Then, PCA is applied to reduce the dimensionality of the data to fewer than 1,000 features. Finally, an SVM or SVR provides the final age estimate. 

The first SFA architecture for age estimation was a four-layer HSFA network that was applied to the images directly without prior feature extraction~\citep{EscalanteWiskott-2010}. 
The input images were synthetic and created using special software for 3D-face modeling. However, the complexity of the face model was probably too simple, which allowed linear SFA to achieve good performance, and left open the question of whether SFA could also be successful on real photographs.

Race and gender are two factors that influence the accuracy of age estimation (e.g., \citep{GuoEtAl-2009,LuuEtAl-2009}).
This idea is exploited by the system in~\cite{GuoMu-2010}, where the faces are first classified according to race and gender, and age is then estimated in the particular race/gender group.
%
Other algorithms allow the estimation of age, race and gender simultaneously. 
Guo et al.\ \cite{GuoMu-2011} proposed the use of kernel partial least squares regression (KPLS) on top of the BIF features. 


More recently, various methods based on canonical correlation analysis (CCA) have been proposed by Guo et al.\ \cite{GuoMu-2014}, particularly regularized kernel CCA (rKCCA), on top of BIF features in a framework for the joint estimation of age, race and gender.

The use of a multi-scale convolutional neural network (MCNN) trained on 23 48$\times$48 image patches has been proposed in \cite{YiEtAl-2014}. Each patch has one out of four different scales and is centered on a particular facial landmark. 



\subsection{Image database and image pre-processing}
\label{sec:database}
\label{sec:image_normalization}
The MORPH-II database (i.e.\ MORPH, Album 2)~\citep{RicanekTesafaye-2006} is a large database available for a symbolic fee and suitable for age estimation.
It contains \num{55134} images of about \num{13000} different persons with ages ranging from 16 to 77 years. The images were taken under partially controlled conditions (e.g.\ frontal pose, absence of glasses, good image quality, absence of strong shadows), and include variations in head pose (e.g.\ tilt angle) and expression.
The database includes annotations stating the age of the persons, their gender (M or F), ``race'': ``black'' (B), ``white'' (W), ``asian'' (A), ``hispanic'' (H), and ``other'' (O), and the coordinates of the eyes. 
The procedure used to assign the race label is unknown to us.
Most of the images are of black (77\%) or white races (19\%), probably making it more difficult to generalize to other races, such as asian. 
We chose this database because of its large number of images. 

We follow the evaluation method used in \cite{GuoMu-2014} and many other works. The input images are partitioned in 3 disjoint sets $S_1$ and $S_2$ of \num{10530} images, and $S_3$ of \num{34074} images. The racial and gender composition of $S_1$ and $S_2$ is the same: they have about 3 times more images of males than females and the same number of white and black people. Other races are omitted.
More exactly, $|MB|=|MW|=\num{3980}$, $|FB|=|FW|=\num{1285}$. The remaining images constitute the set $S_3$, which is composed as follows: $|MB|=\num{28872}$, $|FB|=\num{3187}$, $|MW|=1$, $|FW|=28$, $|MA|=141$, $|MH|=\num{1667}$, $|MO|=44$, $|FA|=13$, $|FH|=102$ and $|FO|=\num{19}$. 
Training and testing are done twice, using either $S_1$ and $S_1\ttxt{-test} \eqdef S_2 + S_3$ or $S_2$ and $S_2\ttxt{-test} \eqdef S_1 + S_3$.

We pre-process the input images in two steps: pose normalization and face sampling (Figure~\ref{fig.input_images}).
The pose-normalization step fixes the position of the eyes ensuring that: 
a) the eye line is horizontal, b) the inter-eye distance is constant, and c) the output resolution is 256$\times$260 pixels. 
After pose normalization, face sampling keeps the head area only, enhances the contrast, and scales down the image to 96$\times$96 pixels. 
Typically the chin, forehead, and some hair are visible in the resulting images. 


We define a DR-dataset to train HiGSFA (or the DR algorithm), an S-dataset to train the supervised step on top of HiGSFA (a Gaussian classifier), and a T-dataset for testing.
The DR and S-datasets are created with the training images (either $S_1$ or $S_2$), and the T-dataset with the corresponding test images, either $S_1\ttxt{-test}$ or $S_2\ttxt{-test}$.

The images of the DR and S-datasets go through a distortion step during face sampling, which include a small random translation of max $\pm 1.4$ pixels, a rotation of max $\pm 2$ degrees, and a rescaling of $\pm 4\%$, as well as small fluctuations in the average color and contrast.
The exact transformations are distributed uniformly in their corresponding intervals.
Although the distortions are frequently imperceptible, they teach HiGSFA to become invariant to small errors during image normalization and are necessary due to its feature specificity to improve generalization to test data.
Other algorithms that use particular structures (e.g., convolutional layers, max pooling) or BIF features are mostly invariant to such small transformations by construction (e.g., \citep{GuoMu-2014}).

Distortions allow us to increase the number of images used for training.  
The images of the DR-dataset are distorted 22 times with different random distortions and those of the S-dataset 3 times, resulting in \num{231660} and \num{31590} images, respectively. The images of the T-dataset are not distorted and used only once.



\subsection{Efficient training graphs for learning multiple-labels}
\label{sec:graph_multilabels}
GSFA would be less attractive without its efficient training graphs, which include a clustered graph (classification) and a serial graph (regression) with a training complexity of $\bigO(NI^2+I^3)$.
However, up to now efficient graphs have only been defined for a single (categorical or numerical) label. 
We briefly recall the clustered and serial graphs~\citep{EscalanteWiskott-2013b}, and then propose an efficient graph for 3 labels. 

\paragraph{Clustered training graph}
This graph generates features useful for classification that are equivalent of those of FDA, see~\citep{KlampflMaass:09b} (also compare~\citep{Berkes05} and~\citep{Berkes-2005a}), and is illustrated in Figure~\ref{fig.clustered_gender}. The optimization problem associated with this graph explicitly demands that samples from the same class should be typically mapped to similar outputs.
If $C$ is the number of classes, $C-1$ output features can be extracted and given to a standard classifier to compute the final class estimate.

\makeatletter%
\if@twocolumn%
	\newcommand{\mysizeclustered}{0.70\columnwidth}
\else
	\newcommand{\mysizeclustered}{0.42\columnwidth}
\fi
\makeatother
\begin{figure}[hbt!]
\begin{small}
\begin{center}
\includegraphics[width=\mysizeclustered]{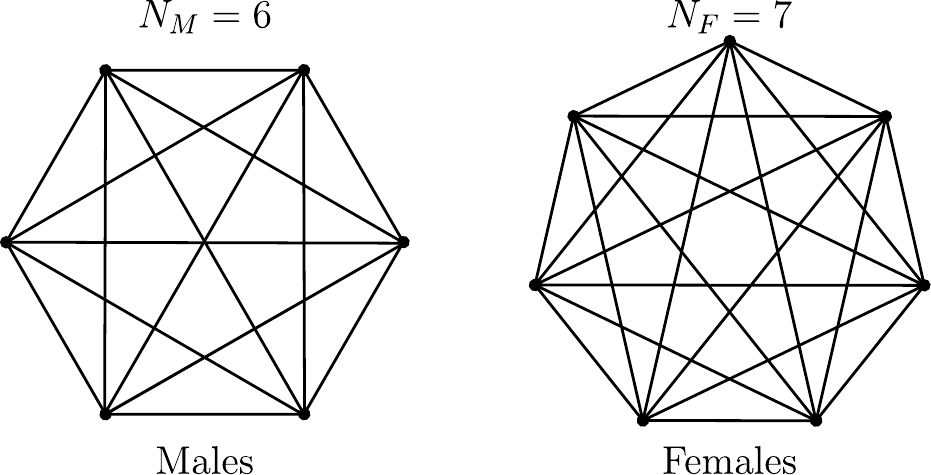}
\caption{Illustration of a \emph{clustered} training graph for gender classification with 6 images of males and 7 of females. Each vertex represents an image and edges represent transitions.
Pairs of images $\vect{x}(n)$ and $\vect{x}(n')$, with $n \neq n'$, that belong to the same class $s$ (either $F$ or $M$) are connected with an edge weight $\e_{n,n'} = 1/(N_s-1)$, where $N_s$ is the number of images in class $s$. This results in 2 fully connected subgraphs. Images of different genders are not connected. The weight of all vertices is equal to one. For the actual experiments, we use $N_F = \num{2570}r$ and $N_M = \num{7960}r$, for $r\eqdef 22$. The graph for race (B and W) is similar with $N_B = N_W = \num{5265}r$.
}
\label{fig.clustered_gender}
\end{center}
\end{small}
\end{figure}

\paragraph{Serial training graph}
\label{sec.serial}
This graph has consistently given good results for different regression problems. Other efficient training graphs for regression are the sliding window and mixed graphs~\citep{EscalanteWiskott-2013b}. After GSFA, a complementary explicit regression step on a few features solves the original regression problem.


The serial graph is constructed by ordering the samples by increasing label. Then, the samples are partitioned into $L$ groups of size $N_g = N/L$. Each group has a representative label $\in \{ \ell_1, \ldots, \ell_{L} \}$, where $\ell_1 < \ell_2 < \cdots < \ell_{L}$, see Figure~\ref{fig.serial_training_graph}.
Edges connect all pairs of samples from two consecutive groups with representative labels ($\ell_{l}$ and $\ell_{l+1}$). Thus, all connections are inter-group, no intra-group connections are present.
Notice that since any two vertices of the same group are adjacent to exactly the same neighbors, they are likely to be mapped to similar outputs by GSFA.

\makeatletter%
\if@twocolumn%
	\newcommand{\mysizeserial}{0.75\columnwidth}
\else
	\newcommand{\mysizeserial}{0.50\columnwidth}
\fi
\makeatother
\begin{figure}[ht!]
\begin{small}
\begin{center}
\includegraphics[width=\mysizeserial]{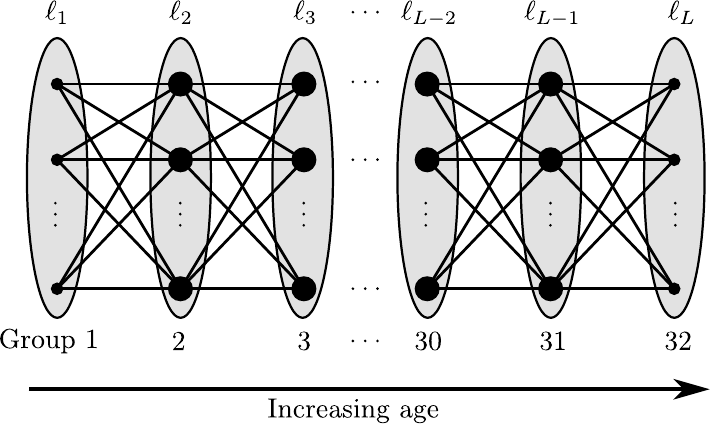} 
\caption{Illustration of a serial training graph for age estimation. The training images are first ordered by increasing age and then grouped into $L=32$ groups of $N_g = \num{7238}$ samples each. 
Each dot represents an image, edges represent connections, and ovals represent the groups. 
The images of the first and last group have weight $1$ and the remaining images have weight $2$ (image weights represented by smaller/bigger dots). The weight of all edges is $1$.
}
\label{fig.serial_training_graph}
\end{center}
\end{small}
\end{figure}


The theory of unrestricted GSFA predicts that mapping the slowest feature may suffice to solve the regression problem accurately (except for a discretization error). This requires that the slowest hidden parameter of the data is a strictly monotonic function of the label. In practice, however, mapping a few slow features frequently gives better results. Different approaches for implementing this mapping have been proposed~\citep{EscalanteWiskott-2013b}. 

\paragraph{Efficient graph for age, race and gender estimation}
Based on a recent analysis of the optimal free responses of GSFA~\cite{EscalanteWiskott-2015a} (i.e., the slowest features possible when the feature space is unlimited), we propose here the combination of $K$ efficient graphs that learn one label each into a single graph that learns $K$ labels. To compute the combined graph, one must only add the vertex and edge weights. 
For the approach to be mathematically sound, we require 3 conditions: (1) all graphs have the same samples and are consistent, (2) all graphs have the same (or proportional) node weights, and (3) optimal free responses that are slow in one graph ($\Delta < 2.0$) should not be fast ($\Delta > 2.0$) in any other graph. These requirements guarantee that the slowest optimal free responses of the combined graph span the slowest optimal free responses of the original graphs.

For example, one can combine a clustered graph for gender (M or F) estimation and another for race (B or W). The first two features of the resulting graph are then enough for gender \emph{and} race classification.
Alternatively, one could create a clustered graph with four classes (MB, MW, FB, FW), so that 3 features would be needed for classification instead of 2.
However, this would be impractical for larger numbers of classes. For example, if the original numbers of classes were $C_1=10$ and $C_2=12$, one would need to extract $C_1 C_2 - 1 = 119$ features, whereas in the proposed graph, one would only need to preserve $(C_1-1) + (C_2 -1) = 20$ features.

To learn the age, race and gender labels, we propose a graph $G_3$ that combines a serial graph for age estimation, a clustered graph for gender, and a clustered graph for race. The serial graph has node weights not quite the same as in the clustered graphs, but this might not affect the accuracy of the combined graph significantly. For comparison, we also use a serial graph $G_1$ that only learns age.

We use the first 4 to 7 features extracted from the S-dataset to train 3 separate Gaussian classifiers (GC).  
For race only 2 classes are considered (B and W), and for gender only M and F.
For age, the images are partitioned in 39 classes of increasing ages. The classes have average ages of  $\{16.6, 17.6, 18.4, \dots, 52.8, 57.8\} $ years.
To compute these average ages, as well as to order the samples by age in the serial graph, we use the age of the persons with a day resolution (e.g., $25.216$ years). However, for the evaluation, we use integer ground-truth labels and integer estimates.


After the GC has been trained, the final age estimation (on the T-dataset) is computed using class membership probabilities.
Let $\Prob(C_{\tilde{\ell}_l} | \vect{y} )$ be the estimated class probability that the input sample $\vect{x}$ with feature representation $\vect{y}= \vect{g}(\vect{x})$ belongs to the group with average label $\tilde{\ell}_l$.
Then, the estimated age is
\begin{equation} \label{eq:regression}
\ell \; \eqdef \; \sum_{l=1}^{39} \tilde{\ell}_l \cdot \Prob(C_{\tilde{\ell}_l} | \vect{y}) \, ,
\end{equation}
where a final step keeps the integer part of $\ell$. Equation (\ref{eq:regression}) is particularly suited to minimize the root mean squared error (RMSE). Although it incurs in an error due to the discretization of the labels, the soft nature of the estimation has provided good accuracy and robustness.

\subsection{Evaluated algorithms}
Besides evaluating HiGSFA, we resort to HGSFA, PCA, and state-of-the-art age-estimation algorithms for comparison purposes.
iGSFA and GSFA are not used directly, but HiGSFA and HGSFA are used to take advantage of the benefits of hierarchical processing. 



The structure of the HiGSFA and HGSFA networks is described in Table~\ref{tab:Networks}.
In both networks, the nodes are composed only of iGSFA or GSFA, with various types of nonlinear expansion functions.
Only in the first layer, PCA is applied to the pixel data prior to iGSFA/GSFA, preserving 20 out of 36 principal components. 
To scale the slow features we use the sensitivity method of Section~\ref{sec:mixing_feats}.
The hyper-parameters have been hand-tuned to achieve best accuracy on age estimation using educated guesses, random sets $S_1$, $S_2$ and $S_3$ different to those used for the evaluation, and fewer image multiplicities to speed up the process.

The HGSFA/HiGSFA networks differ in several aspects from SFA networks used in the literature (e.g., \citep{FSW07}). For example, to improve feature specificity at the lowest layers, no weight sharing was used.
Moreover, the input to the nodes (fan-in) originates mostly from the output of 3 nodes in the preceding layer (3$\times$1 or 1$\times$3). Such a small fan-in reduces the computational cost because the input dimensionality is minimized. It also results in networks with 10 layers, potentiating the accumulation of nonlinearity across the network.

The expansion functions are a mixture of different nonlinear functions on subsets of the input vectors, including: (1) The identity function $\tsys{I}(\vect{x}) = \vect{x}$. (2) The quadratic terms $\tsys{QT}(\vect{x}) \eqdef \{ x_i x_j \}_{i,j=1}^N$. (3) A normalized version of $\tsys{QT}$: $\tsys{QN}(\vect{x}) \eqdef \{ \frac{1}{1+||\vect{x}||^2} x_i x_j \}_{i,j=1}^N$. (4) The terms $\tsys{0.8ET}(\vect{x}) \eqdef \{ |x_i|^{0.8} \}_{i=1}^N$, which are useful to improve generalization and resistance against outliers \citep{EscalanteWiskott-2011}.
(5) The function $\tsys{max2}(\vect{x}) \eqdef \{ \max(x_i,x_{i+1}) \}_{i=1}^{N-1}$. We propose this function inspired by the state-of-the-art algorithms on age estimation, which include max pooling or some variation of it. For example, the nonlinear expansion of the first layer of the HiGSFA network is $\tsys{I}(x_1, \dots, x_{18}) \, | \allowbreak \tsys{0.8ET}(x_1, \dots, x_{15}) \,|\, \tsys{max2}(x_1, \allowbreak \dots, \allowbreak x_{17}) \,|\, \allowbreak \tsys{QT}(x_1, \dots, x_{10})$, where $|$ indicates vector concatenation. 
The details of the expansions used in the remaining layers are available upon request.

\begin{centering} \begin{table}[htb!] \begin{center} \centering \footnotesize \begin{tabular}{cccccc}
\toprule
\multirow{2}{*}{Layer} & \multirow{2}{*}{size} & node    & \multirow{2}{*}{stride} & \multicolumn{2}{c}{output dim. per node}  \\
\cmidrule(r){5-6}
                   &                       & fan-in      &                     & HGSFA  & HiGSFA  \\
\midrule 
0                  & 96$\times$96 pixels   & ---         & ---        & --- & ---   \\
1                  & 31$\times$31 nodes    & 6$\times$6  & 3$\times$3 & 14  & 18   \\
2                  & 15$\times$31 nodes    & 3$\times$1  & 2$\times$1 & 20  & 27   \\
3                  & 15$\times$15 nodes    & 1$\times$3  & 1$\times$2 & 27  & 37   \\
4                  &  7$\times$15 nodes    & 3$\times$1  & 2$\times$1 & 49  & 66   \\
5                  &  7$\times$7 nodes     & 1$\times$3  & 1$\times$2 & 60  & 79   \\
6                  &  3$\times$7 nodes     & 3$\times$1  & 2$\times$1 & 61  & 88   \\
7                  &  3$\times$3 nodes     & 1$\times$3  & 1$\times$2 & 65  & 88   \\
8                  &  1$\times$3 nodes     & 3$\times$1  & 1$\times$1 & 65  & 93   \\
9                  &  1$\times$1 nodes     & 1$\times$3  & ---        & 66  & 95  \\
10                 &  1$\times$1 nodes     & 1$\times$1  & ---        & 75  & 75   \\
\bottomrule 
\end{tabular} \end{center} \caption{Description of the HiGSFA and HGSFA networks. 
The two networks have the same number of nodes and general structure, but they differ in the type of nodes and in the number of features these preserve.
Layer 0 denotes the input image, whereas layer 10 is the top node.
The size of the slow part in layers 1 and 2 does not depend on $\Delta_T$ but is fixed to 3 or 4 features, resp.
}
\label{tab:Networks} \end{table} \end{centering} 

The parameter $\Delta_T$ of layers 3 to 10 was set to \num{1.96}. $\Delta_T$ was not used in layers 1 and 2, where the number of slow features was fixed to 3 and 4, resp. The number of features given to the supervised algorithm, shown in Table~\ref{tab:num_feats_GC}, was tuned for each DR algorithm and supervised problem.
\begin{centering} \begin{table}[htb!] \begin{center} \centering \footnotesize \begin{tabular}{cccc}
\toprule
Algorithm & Age & Race & Gender \\
\midrule
HiGSFA ($G_3$) &  5    &  6   &  4 \\
HGSFA ($G_3$)  &  5    &  5   &  7 \\
PCA     & 54    & 54   & 60 \\
\bottomrule
\end{tabular} \end{center} \caption{Number of output features given to the supervised step (a Gaussian classifier).}
\label{tab:num_feats_GC} \end{table} \end{centering}

Since the data dimensionality allows it, we used PCA directly (contrary to hierarchical PCA) to provide more accurate principal components and smaller reconstruction errors.

\subsection{Experimental results}
We show now the results of HiGSFA, HGSFA and PCA on the three evaluation criteria. 
Results are reported as $a \pm b$, where $a$ is the average over the test images ($S_1\ttxt{-test}$ and $S_2\ttxt{-test}$), and $b$ is the standard error of the mean (i.e., half the absolute difference).

\subsubsection{Feature slowness}
The weighted $\Delta$ values of GSFA (Equation \ref{eq:wobj}), which we denote here as $\Delta^{\ttxt{DR},G_3}_j$, depend on the graph $G_3$, which in turn depends on the training data and their labels.
To measure slowness (or rather fastness) of test data, we compute standard $\Delta$ values on the images ordered by increasing age label,
$\Delta^{\ttxt{T,lin}}_j \eqdef \frac{1}{N-1}\sum_n (y_j(n+1)-y_j(n))^2$, which is equivalent to using a linear graph.
In all cases, the scale of the features is normalized to variance 1 before computing the $\Delta$ values to prevent alterations due to the feature scaling method. 

Table~\ref{tab:DeltaValues} shows $\Delta^{\ttxt{DR},G_3}_{1,2,3}$ (resp.  $\Delta^{\ttxt{T,lin}}_{1,2,3}$), that is, the $\Delta$ values of the three slowest features extracted from the DR-dataset (resp. T-dataset) using the graph $G_3$ (resp. a linear graph).

\begin{centering} \begin{table}[htb!] \begin{center} \centering \footnotesize \begin{tabular}{cccc}
\toprule
                             &          PCA & HGSFA ($G_3$) & HiGSFA ($G_3$)\\
\midrule
$\Delta_1^{\ttxt{DR},G_3}$  & --   & \num{1.23} & \num{1.17} \\
$\Delta_2^{\ttxt{DR},G_3}$  & --   & \num{1.46} & \num{1.38} \\
$\Delta_3^{\ttxt{DR},G_3}$  & --   & \num{1.56} & \num{1.53} \\
\midrule
$\Delta_1^\ttxt{T,lin}$   & \num{1.99} & \num{0.45} & \num{0.38} \\
$\Delta_2^\ttxt{T,lin}$   & \num{1.93} & \num{1.12} & \num{0.99} \\
$\Delta_3^\ttxt{T,lin}$   & \num{1.99} & \num{1.90} & \num{1.90} \\
\bottomrule
\end{tabular} \end{center} 
\caption{Average delta values of the first three features extracted by PCA, HGSFA and HiGSFA on training and test data (the smaller the better). 
The first feature extracted is the most stable according to the age-ordered linear graph, which indicates that this is the main feature coding age. 
For comparison, the $\Delta$ value of unit-variance i.i.d. noise is $2.0$.
}
\label{tab:DeltaValues} \end{table} \end{centering} 

Table~\ref{tab:DeltaValues} shows that HiGSFA outperforms HGSFA in slowness maximization.
The $\Delta^\ttxt{T,lin}$ values of the PCA features are larger, which is not surprising, because PCA does not optimize for slowness.
Since $\Delta^{\ttxt{DR},G_3}$ and $\Delta^\ttxt{T,lin}$ are computed from different graphs, they should not be compared with each other. 
$\Delta^\ttxt{T,lin}$ considers transitions between images with the same or very similar ages but arbitrary race and gender.
$\Delta^{\ttxt{DR},G_3}$ only considers transitions between images having at least one of a) the same gender, b) the same race, or c) different but consecutive age groups. 


\subsubsection{Age estimation error}
\label{sec:results_age_estimation}
Some real-life applications only need a coarse categorization of age in broad age groups.
However, other applications benefit from a more precise estimation, making it convenient to treat age estimation as a regression problem requiring a concrete numerical estimation, usually expressed as an integer number of years. 

We use three metrics to measure age estimation accuracy:
(1) The mean absolute error (MAE), see \citep{GengZhouSmith-2007}, which is the most frequent metric for age estimation,
(2) the root mean squared error (RMSE), which is a common loss function for regression in Machine Learning, and (3) cumulative scores (CSs), see \citep{GengZhouSmith-2007}, which indicate the fraction of the images that have an estimation error below a given threshold. 
For instance, $\tvar{CS}(5)$ is the fraction of estimates (e.g., expressed as a percentage) having an error of at most 5 years w.r.t.\ the real age. 
We include CSs at various thresholds, facilitating future comparisons with other methods.
Although the RMSE is sensitive to outliers and has almost not been used in the literature on age estimation, some applications might benefit from its stronger penalization of larger estimation errors. The accuracies are summarized in Table~\ref{tab:MAE}.


\begin{centering} \begin{table*}[htb!] \begin{center} \centering \footnotesize \begin{tabular}{ccccc}
\toprule
Algorithm & Age (MAE) & Age (RMSE) & Race (CR) & Gender (CR) \\
\midrule
BIF+3Step~\citep{GuoMu-2010} & $4.45 \pm 0.01$ & --- & $98.80\% \pm 0.04$ & $97.84\% \pm 0.16$\\
BIF+KPLS~\citep{GuoMu-2011} & $4.18 \pm 0.03$ & --- & $98.85\% \pm 0.05$ & $98.20\% \pm 0.00$\\
BIF+rKCCA~\citep{GuoMu-2014} & $3.98 \pm 0.03$ & --- & $99.00\% \pm 0.00$ & $\bf{98.45\% \pm 0.05}$\\
BIF+rKCCA+SVM~\citep{GuoMu-2014} & $3.92 \pm 0.02$ & --- & --- & ---\\
baseline CNN~\citep{YiEtAl-2014} & $4.60 \pm 0.05$  & --- & --- & ---\\
MCNN$^\star$ no align~\citep{YiEtAl-2014} & $3.79 \pm 0.09$  & --- & --- & ---\\
MCNN$^\star$ only age~\citep{YiEtAl-2014} & $3.63 \pm 0.00$  & --- & --- & ---\\
MCNN$^\star$~\citep{YiEtAl-2014} & $3.63 \pm 0.09$  & --- & $98.6\% \pm 0.05$ & $97.9\% \pm 0.1$\\

\midrule
PCA        (control)  & $6.804 \pm 0.007$  & $8.888 \pm 0.000$ & $96.75\% \pm 0.06$ & $91.54\% \pm 0.24$ \\
HGSFA ($G_3$)   (control)  & $3.921 \pm 0.018$  & $5.148 \pm 0.049$ & $98.60\% \pm 0.08$ & $96.40\% \pm 0.12$ \\
HiGSFA ($G_1$) (control) & $3.605 \pm 0.001$ & $4.690 \pm 0.000$ & --- & ---  \\
HiGSFA ($G_3$) (proposed) & $\bm{3.497 \pm 0.008}$ & $\bm{4.583 \pm 0.000}$ & $\bm{99.15\% \pm 0.01}$ & $97.70\% \pm 0.01$  \\
\midrule
Chance level &  $9.33$  & $10.95$ & $87.58\%$ & $86.73\%$ \\
\bottomrule
\end{tabular} \end{center} \caption{Accuracy in years of state-of-the-art algorithms for age estimation on the MORPH-II database (test data). Classification rates (CR) for race and gender estimation are also provided. The chance level is the best possible performance when the estimation is constant. $^\star$A mistake in the evaluation protocol of MCNN made their training and test data not disjoint, thus the actual accuracy might differ, see \protect\url{http://www.cbsr.ia.ac.cn/users/dyi/agr.html}.}
\label{tab:MAE} \end{table*} \end{centering}


The MAE of HGSFA was $3.921$ years, which is better than that of BIF+3Step, BIF+KPLS and BIF+rKCCA, similar to BIF+rKCCA+SVM, and worse than the MCNNs (except the baseline).
However, the MAE of HiGSFA is only $3.497$ years, which is better than all previous algorithms we have found published. In contrast, with an MAE of $6.804$ years PCA has the largest MAE.
Detailed cumulative scores for HiGSFA and HGSFA are provided in Table~\ref{tab:CumulativeScores}. 

The RMSE of HGSFA on test data is $5.148$ years, HiGSFA yielded an RMSE of $4.583$ years, and PCA an RMSE of $8.888$ years. The RMSE of other approaches does not seem to be available. 

The poor accuracy of PCA for age estimation is not surprising, because principal components might lose wrinkles, skin imperfections, and other information that could reveal age in adult persons, and it might also be explained because principal components are too unstructured to be properly untangled by the Soft GC method, in contrast to slow features, which have a very specific and simple structure.

\begin{centering} \begin{table*}[htb!] \begin{center} \centering \footnotesize \begin{tabular}{cccccccccccccccccccc}
\toprule
Algorithm & cs(0) & cs(1) & cs(2) & cs(3) & cs(4) & cs(5) & cs(6) & cs(7) & cs(8) & cs(9) & cs(10)  \\
\midrule
HGSFA     & 8.80 & 26.42 & 42.41 & 55.80 & 66.38 & 74.86 & 81.31 & 86.37 & 90.12 & 92.93 & 94.96 \\
HiGSFA   & 9.87 & 29.16 & 46.23 & 60.14 & 71.04 & 79.56 & 85.70 & 90.04 & 93.12 & 95.45 & 96.92 \\
\bottomrule
\end{tabular}\\
\vspace{0.15cm}
\begin{tabular}{cccccccccccccccccccc}
\toprule
            &  cs(11) &  cs(12) & cs(14) & cs(16) & cs(18) & cs(20) & cs(22) & cs(24) & cs(26) & cs(28) & cs(30)\\
\midrule
HGSFA       & 96.43 & 97.42 & 98.67 & 99.34 & 99.69 & 99.85 & 99.92 & 99.97 & 99.97 & 99.98 & 99.99 \\
HiGSFA     & 97.91 & 98.56 & 99.34 & 99.70 & 99.84 & 99.91 & 99.94 & 99.96 & 99.97 & 99.98 & 99.99 \\
\bottomrule
\end{tabular} \end{center} \caption{Percentile cumulative scores (the larger the better) for various maximum allowed errors ranging from 0 to 30 years.}
\label{tab:CumulativeScores} \end{table*} \end{centering} 

The behavior of the estimation errors as a function of the real age is plotted in Figure~\ref{fig.maes}. On average, older persons are estimated much younger than they really are. This is in part due to the small number of older persons in the database, and because the oldest class used in the supervised step has an average of about 58 years, making this the largest age that can be estimated. 
The MAE is surprisingly low for persons below 45 years. For 19-year-old persons, the MAE is only \num{2.253} years.
\makeatletter%
\if@twocolumn%
	\newcommand{\mysizemaes}{0.80\columnwidth}
\else
	\newcommand{\mysizemaes}{0.60\columnwidth}
\fi
\makeatother
\begin{figure}[ht!]
\begin{small}
\begin{center}
\includegraphics[width=\mysizemaes]{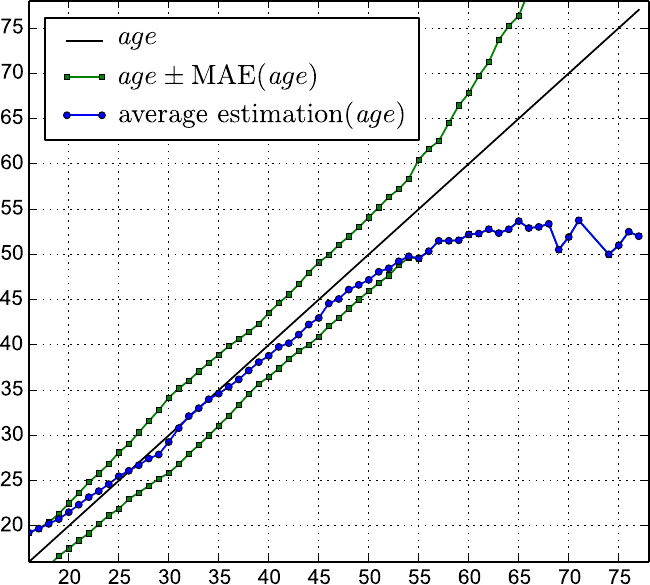}
\caption{
The average age estimate is plotted as a function of the real age. The MAE is also computed as a function of the real age and plotted as $\tvar{age} \pm \ttxt{MAE}(\tvar{age})$.
}
\label{fig.maes}
\end{center}
\end{small}
\end{figure}

\subsubsection{Reconstruction error}
A reconstruction error provides an indication of how much information the output features retain from the original input. In order to compute it, we assume a linear global model for input reconstruction.

Let $\vect{X}$ be the input data and $\vect{Y}$ the corresponding set of extracted features.
A matrix $\vect{D}$ and a vector $\vect{c}$ are learned from the DR-dataset using linear regression such that $\vect{\hat{X}} \eqdef \vect{D} \vect{Y} + \vect{c}\vect{1}^T$ approximates $\vect{X}$ as closely as possible, where $\vect{1}$ is a vector $N$ of ones.
$\vect{\hat{X}}$ is a matrix containing the reconstructed samples (i.e.\ $\vect{\hat{x}_n} \eqdef \vect{D} \vect{y_n} + \vect{c}$ is the reconstruction of the input $\vect{x_n}$ given its feature representation $\vect{y_n}$).
Figure~\ref{fig.input_images} provides examples of face reconstructions from different features.


Since the model is linear and global 
the output features are mapped to the input linearly. 
For PCA this gives the same result as the usual multiplication with the transposed projection matrix plus image average.
An alternative (local) approach for HiGSFA would be to use the pseudo-inversion algorithm to perform reconstruction from the top of the network to the bottom, one node at a time.

The normalized reconstruction error, computed from the T-dataset, is then defined as 
\begin{equation}
e_\ttxt{rec} \eqdef \frac{\sum_{n=1}^{N} || (\vect{x_n} - \vect{\hat{x}_n}) || ^ 2 }{\sum_{n=1}^{N} || (\vect{x_n} - \bar{\vect{x}}) || ^ 2 } \, , 
\end{equation}
which is the ratio between the energy of the reconstruction error and the variance of the test data except for a factor $N/N-1$. 

\begin{centering} \begin{table}[hbt!] \begin{center} \centering \footnotesize \begin{tabular}{ccccc}
\toprule
               & Chance level & HGSFA & HiGSFA & PCA     \\
\midrule
$e_\ttxt{rec}$ & 1.0          & 0.818 & 0.338 & \bf{0.201} \\
\bottomrule 
\end{tabular} \end{center} 
\caption{Reconstruction errors on test data using 75 features and various algorithms.}
\label{tab:reconstruction_error}
\end{table} \end{centering} 

The reconstruction errors of HGSFA, HiGSFA and PCA using 75 features are given in Table~\ref{tab:reconstruction_error}.
The largest reconstruction error results from the constant reconstruction $\vect{\bar{x}}$ (chance level). As expected, HGSFA does slightly better than chance level, but worse than HiGSFA, which is closer to PCA. PCA yields the best possible features for the given linear global reconstruction method, and is better than HiGSFA by 0.127.
For HiGSFA, from the 75 output features, 8 of them are slow features (slow part), and the remaining 67 are reconstructive. 
Using 67 features, PCA yields a reconstruction error of \num{0.211}.

\subsubsection{HiGSFA network with HGSFA hyper-parameters}
In the experiments above, the hyper-parameters of the HiGSFA and HGSFA networks (e.g., nonlinear expansion functions, output dimensionalities) were tuned separately.
To verify that the performance of HiGSFA is better than that of HGSFA not simply due to different hyper-parameters, we evaluated the performance of an HiGSFA network using the hyper-parameters of the HGSFA network (the only difference is the use of iGSFA nodes instead of GSFA nodes).
The hyper-parameter $\Delta_T$, not present in HGSFA, was set as in the tuned HiGSFA network.
As expected, the performance of HiGSFA was affected: The MAE increased to $3.72$ years, and the RMSE increased to $4.80$ years. The reconstruction error improved slightly to $0.322$. 
Although the suboptimal hyper-parameters affected HiGSFA, it was still clearly superior to HGSFA.

\subsubsection{Sensitivity to the delta threshold $\Delta_T$}
We have evaluated the influence of $\Delta_T$ on estimation accuracy and numerical stability, by testing different values for $\Delta_T$.
%
For simplicity, we used the same $\Delta_T$ from layers 3 to 10 in this experiment ($\Delta_T$ is not used in layers 1 and 2, where the number of features in the slow part is constant, 3 and 4 features, respectively).
The performance of the algorithm as a function of $\Delta_T$ is shown in Table~\ref{tab:DeltaThresholds}. The $\Delta_T$ yielding minimum MAE and used in the optimized architecture is 1.96. 

\begin{centering} \begin{table}[htb!] \begin{center} \centering \footnotesize \begin{tabular}{cccccccc}
\toprule
\multirow{2}{*}{$\Delta_T$} & Age & Age & \multirow{2}{*}{$e_\ttxt{rec}$} & Race      & Gender        & \#Features \\
           &  MAE        &   RMSE       &          & CR      & CR    & L3 \\
\midrule 
1.92       & 3.506    & 4.583    & 0.330    &  99.15    & 97.58     & 2.87     \\
1.94       & 3.499    & 4.583    & 0.333    &  99.15    &  97.64    & 3.22     \\
1.96       & 3.497    & 4.583    & 0.338    &  99.15    & 97.00     & 3.66     \\ 
1.98       & 3.530    & 4.637    & 0.359    &  99.09    & 94.72     & 4.14     \\
\bottomrule
\end{tabular} \end{center} \caption{Performance of HiGSFA on the MORPH-II database using different $\Delta_T$ (default value is $\Delta_T=1.96$).
The results reported are the age estimation errors (MAE and RMSE), the reconstruction error $e_\ttxt{rec}$, the percentile classification rate for race and gender, and the average number of features with $\Delta < \Delta_T$ in the nodes of the third HiGSFA layer. 
All error measures were computed on test data.
}
\label{tab:DeltaThresholds} \end{table} \end{centering} 

The average number of slow features in the third layer changed moderately depending on  $\Delta_T$, ranging from $2.87$ to $4.14$ features, and the differences in the final error measures were small.
This shows that the parameter $\Delta_T$ is not critical and can be easily tuned.


\subsubsection{Evaluation on the FG-NET database}
\begin{sloppypar}
We investigated the capability of HiGSFA to generalize to a different database. We used the HiGSFA $(G_3)$ network trained with images of the MORPH-II database (either with the set $S_1$ or $S_2$), and tested it using images of the FG-NET database~\cite{FGNet-Database}. The test images were taken under uncontrolled conditions (e.g., many are not frontal) and we excluded those out of the original age range of 16 to 77 years.
\end{sloppypar}

For age estimation, the MAE was 7.32 $\pm$ 0.08 years and the RMSE 9.51 $\pm$ 0.13 years (using 4 features for the supervised step). 
For gender and race estimation, the classification rates (5 features) are 80.85\% $\pm$ 0.95\% and 89.24\% $\pm$ 1.06\%, resp. We assumed white race for all the test persons.

The most comparable cross-database experiment known to us is a system trained on a large database of images from the internet and tested on FG-Net~\cite{NiEtAl-2011}. If the same age range as above is used, the MAE is approximately 8.29 years. 

\subsubsection{Alternative evaluation protocol}
The protocol used to create the training and test data (based on the sets $S_1$, $S_2$ and $S_3$)~\cite{GuoMu-2014} has been frequently used for age estimation. 
However, it has a few disadvantages.
Thus, we also adopt the ``leave one person out'' (LOPO)~\citep{ChoiEtAl-2011,HuertaEtAl-2015} protocol. For efficiency reasons, we leave 2000 persons out of the training set instead of one. 
That is, the test data are created with the images of \num{2000} persons chosen at random (about \num{8000} images), whereas the training data is simply the remaining images (about \num{45000} images).

This protocol has the following properties in contrast to the one used in~\cite{GuoMu-2014}:
(a) The distribution of the training and test images is the same on average: $|M|=85\%$, $|F|=15\%$, $|B|=77\%$, $|W|=19\%$, $|H|=3\%$, $|A|<1\%$, $|O|<1\%$. 
This is a basic assumption in machine learning.  
In contrast, the original protocol has: $|M|=76\%$, $|F|=24\%$, $|B|=50\%$, $|W|=50\%$, $|O|=|A|=|H|=0\%$ for training, and $|M|=87\%$, $|F|=13\%$, $|B|=84\%$, $|W|=12\%$, $|H|=4\%$, $|A|<1\%$, and $|O|<1\%$ for testing.
(b) The number of training images available is 4 times larger (about 45,000 vs 10,530 images), which might improve generalization.
(c) Since realistic applications might involve age estimations from unknown persons, we restrict the test images to persons not appearing in the training data. In \cite{GuoMu-2014}, $44\%$ of the test images belong to persons seen in training.
(d) One can improve evaluation accuracy by repeating the protocol several times (here 5 times). The original protocol is repeated twice.

The alternative protocol has more training images. Thus, we distort each image only 4 times to create the DR-dataset and once to create the S-dataset. Notice that all races appear in the training and test data. 
Therefore, during training we consider two classes, B and a virtual class R=W+A+H+O to preserve a binary clustered graph for race estimation and balance the size of the classes. However, for better comparison, to compute the CR for race only the B and W races are considered. The results are shown in Table~\ref{tab:HiGSFA_mysetup}. 

\begin{centering} \begin{table*}[htb!] \begin{center} \centering \footnotesize \begin{tabular}{cccccc}
\toprule
Algorithm & Age (MAE) & Age (RMSE) & Race (CR) & Gender (CR) \\
\midrule
CNN (5CV)$^{\star\ddagger}$~\cite{HuertaEtAl-2015} & 3.88 & --- & --- & --- \\
BIF+3SVM $^{\dagger\ddagger}$~\citep{HanEtAll-2013} &$4.2 $          & --- & --- & ---\\
\midrule
HiGSFA ($G_3$)    &  $3.511 \pm 0.002$ & $4.626 \pm 0.026$ & \num{98.80}\% $\pm$ \num{0.01} & \num{98.53} \% $\pm$ \num{0.02}  \\
HiGSFA (age only) &  $\bm {3.502 \pm 0.003}$ & $\bm{4.583 \pm 0.000}$ & --- & --- \\
\midrule 
Chance level & $9.16$  & $10.78$ & $87.53\%$ & $80.69\%$ \\
\bottomrule
\end{tabular} \end{center} \caption{Accuracy of HiGSFA on MORPH-II using the alternative protocol. Some results are written in this table as $a \pm b$, where $a$ is the average over 5 runs (test data) and $b$ is the standard error of the mean. $^\star$5-fold cross validation. $^\dagger$Used a larger version of the database but only \num{10001} training images.  
$^\ddagger$As in the alternative protocol, it was ensured that the persons in the training and test data are disjoint. 
}
\label{tab:HiGSFA_mysetup} \end{table*} \end{centering}

\section{Discussion}
\label{sec:discussion}
\begin{sloppypar}
In this article, we propose an extension of HGSFA called hierarchical information-preserving GSFA (HiGSFA) that complements the slowness principle with information preservation improving global slowness, input reconstruction and estimation accuracy for supervised learning problems.
\end{sloppypar}

We analyze the advantages and limitations of HSFA (and HGSFA) networks, particularly the phenomena of unnecessary information loss and poor input reconstruction.
Unnecessary information loss occurs when a node in the network prematurely discards information that would have been useful for slowness maximization in another node higher up in the hierarchy.
Poor input reconstruction refers to the difficulty of approximating an input accurately from its feature representation.
We show that these phenomena are the result of locally optimizing slowness, yielding suboptimal global features.

To address these shortcomings, we improve feature extraction at the local level. The feature vectors computed by the iGSFA nodes of HiGSFA are divided in two parts, a slow and a reconstructive part. 
The features of the slow part follow a slowness optimization goal and are slow features transformed by a linear scaling. 
The features of the reconstructive part follow the principle of information preservation (i.e.\ maximization of mutual information between outputs and labels), which we implement in practice as the minimization of a reconstruction error.
A parameter $\Delta_T \approx 2.0$ balances the lengths of the slow and reconstructive parts, $J$ and $D-J$ features respectively, where $D$ is the output dimensionality and $J$ is the number of features with $\Delta < \Delta_T$.

We also present a new method to combine various efficient training graphs into a single efficient training graph. With the combined graph we can learn several (numerical or categorical) labels simultaneously. 
Its construction is simple; one only has to add the node and edge weights of the graphs, and the only requirements are that the graphs are consistent and have the same or proportional node weights. 

The experimental results show that HiGSFA is better than HGSFA in terms of feature slowness, input reconstruction and age estimation accuracy. Moreover, HiGSFA offers higher accuracies than current state-of-the-art algorithms for age estimation, including approaches based on BIF features and convolutional neural networks. 
The improvement is a reduction by $48.5$ days ($\approx $ 1.5 months) in the average estimation error. This is a relatively small but significant improvement.

In the next sections, we discuss a few conceptual aspects of the proposed approach, the results and future work.

\subsection{The approach}
Information preservation can be guaranteed by preserving the information contained in the data that describes the global slow features. However, we show that one cannot always identify this information at a local level. Therefore, we resort to the reconstruction goal and preserve as much information of the local input as possible, which is likely to also include information relevant to extract the global slow parameters.  

We addressed feature garbling only briefly and are aware that this is a complex problem that needs better formalization. HiGSFA may only partially reduce this problem by the use of reconstructive features, which might be simpler than slow features. However, feature garbling is still present in the slow part.

The features extracted by HiGSFA are better than those of HGSFA quantitatively and qualitatively. 
Even if unlimited training data and computational resources were available, the features extracted by HGSFA do not necessarily converge to those of HiGSFA. 
In this overfitting-free scenario, information loss would only decrease partially in HGSFA, because the main cause of this problem is not overfitting but the local optimization of slowness.

There is a subtle similarity between the overall unsupervised learning of HiGSFA and some techniques for training CNNs (unsupervised pre-training and joint optimization of supervised and unsupervised objectives). These approaches result in more information from the input reaching the top nodes. 

Another way to combine the slowness principle with information preservation is to optimize a single objective function that integrates both criteria, favoring directions that are slow and have a large variance. However, we found in previous experiments that balancing the two criteria is difficult in practice.

HiGSFA inherits from SFA a close connection to unsupervised learning.
Like GSFA can be emulated with SFA (see \citep{EscalanteWiskott-2013b}) HiGSFA can be approximated by training HiSFA unsupervisedly with data generated from a particular Markov chain. This emulation would incur in a small error due to PCA being unaware of sample weights, which could be fixed by using the node weights as weighting factors of the samples during the computation of the covariance matrix by PCA. 


\subsection{Network parameters}
By selecting the network structure appropriately, the computational complexity of HiGSFA (and other hierarchical versions of SFA) is linear w.r.t.\ the number of samples and their dimensionality, offering feasible training times.
Training a single HiGSFA network (\num{231660} images of 96$\times$96 pixels) takes only 10 hours, whereas HiSFA takes about 6 hours (including the time needed for data loading, the supervised step, and the evaluation) on a single computer (24 cores and 128 GB of RAM) without GPU computing.
However, the algorithm can be implemented on GPUs or using distributed processing, because the nodes in a layer may be trained independently. 
For comparison, the system of~\cite{GuoMu-2014} takes 24.5 hours (training and testing).

HiGSFA is more accurate than HGSFA with the same output dimensionalities and network hyper-parameters.
However, HiGSFA yields even higher accuracies if larger output dimensionalities are used than those of the HGSFA network.
This can be explained by various factors: 
(a) In both networks, the input dimensionality of SFA is $I'$ (the expanded dimension), whereas in HiGSFA the input dimensionality of PCA is $I$, where typically $I' \gg I$. Hence, slow features may overfit more than reconstructive features for this setup. 
(b) In HGSFA, the features of all the nodes are attracted to the same optimal free responses, whereas in HiGSFA, the reconstructive part is attracted to the (different for each node) local principal components. Thus, in HGSFA overfitting might accumulate through the layers more than in HiGSFA.
(c) The HGSFA network may benefit less from more input features, because faster features might be noise-like and result in more overfitting without providing much additional information.

In the iGSFA algorithm, the number of slow features $J$ with $\Delta < \Delta_T$ should be smaller than the output dimensionality $D$, so that $D-J$ output features are reconstructive.
Otherwise, the output features would not have a reconstructive part, removing the advantages of the method.
This might happen if too many slow parameters, their mixtures, or higher-frequency harmonics are present in the data. 
One might avoid this problem by setting a smaller $\Delta_T$, by directly controlling the number of the slow features, and by using training graphs with a small number of optimal free responses with $\Delta < 2.0$.


\subsection{Age, gender and race estimation}
We choose the problem of age estimation from adult facial photographs, because it appears to be an ideal problem to test the capabilities of HiGSFA. For age estimation, PCA is not very useful because wrinkles, skin texture and other higher-frequency features are poorly represented. Therefore, it is not obvious and even counter-intuitive that feature slowness improves by incorporating PCs in HiGSFA.
Improvements on feature slowness using other supervised learning problems, such as gender, race or horizontal-position estimation, would be inconclusive because for such problems a few PCs code the discriminative information relatively well.

To estimate age, race and gender simultaneously, we propose a graph $G_3$ that combines three graphs, encoding sensitivity to the particular labels and favoring invariance to any other factor.
The graphs used are a serial graph for age estimation with 32 groups and two clustered graphs with 2 clusters each for race and gender estimation.
The number of features with $\Delta<2.0$ that can be extracted from these graphs is 15, 1 and 1, respectively (due to their particular geometry). 
Since the node weights of these graphs are not proportional exactly, the combined graph has more than 17 optimal responses with $\Delta<2.0$. Still, the first 5 features contain most of the relevant information, and the method works well in practice.

The reconstructive part is not used in the supervised step, only the first 5 slow features of the output.
This shows that HiGSFA and HGSFA concentrate the label information in the first features.
One can actually replace the iGSFA node on the top of the HiGSFA network by a regular GSFA node, so that all features are slow, without affecting the performance.
The superiority in age estimation of HiGSFA over HGSFA is thus not due to the use of principal components in the final supervised step but to the higher quality of the slow features.

%
The performance of HiGSFA for age estimation is the highest reported on the MORPH-II database with an MAE of 3.497 years. Previous state-of-the-art results are an MAE of 3.63 years using a multi-scale CNN \cite{YiEtAl-2014} and 3.92 using BIF+rKCCA+SVM \cite{GuoMu-2014}.

Even though our system performs slightly better than state-of-the-art algorithms, this was not our focus here.
Our goal was to improve on HGSFA. Thus, our claim is that HiGSFA is better than HGSFA regarding feature slowness, input reconstruction and estimation accuracy.

\subsection{Reconstruction from slow features}
The experiments confirm that PCA is more accurate than HiGSFA at reconstruction using 75 features, which was expected because PCA features are optimal when reconstruction is linear. 
In HiGSFA 67 features are reconstructive (8 less than in PCA) and they are computed hierarchically (locally), in contrast to the PCA features, which are global.
Thus, it is encouraging that the gap between PCA and HiGSFA at input reconstruction is moderate.
In turn, HiGSFA is more accurate than HGSFA, because reconstruction is the secondary goal of HiGSFA, whereas HGSFA does not pursue reconstruction. 
The improved reconstruction capability of HiGSFA might facilitate certain applications, such as morphing.

Since the HiGSFA network implements a nonlinear transformation, nonlinear reconstruction algorithms are also reasonable.
Nonlinear reconstruction might provide more accurate reconstructions in theory, but we have not been able to train such type of algorithms well enough to perform better on test data than the simpler global linear reconstruction algorithm.
One algorithm for nonlinear reconstruction minimizes the feature error. However, since the number of dimensions is reduced by the network, one can expect many samples with feature error $e_\ttxt{feat}=0$ (or $e_\ttxt{feat}$ minimal) that differ substantially from any valid input sample, not looking at all like a face. To correct this problem, one might need to consider the input distribution to select an appropriate reconstruction.
Although generative adversarial networks~\cite{GoodfellowEtAl-2014,RadfordEtAl-2015,DentonEtAll-2015} were originally designed to generate random inputs, it might be possible to adapt them to do reconstruction from HiGSFA networks.

\subsection{Future work}
\label{sec:future_work}

Estimation accuracy may be improved by using more complex hierarchical networks, for instance, by increasing the overlap of the receptive fields and using more complex nonlinearities. As usual, more training images might also improve accuracy, which could be approximated by implementing true face-distortion methods and not just simple transformations at the image level.

One key factor for the performance of multi-scale CNN is the use of receptive fields centered at specific facial points (compare with non-aligned receptive fields). 
This idea could also be applied to HiGSFA, and might particularly boost generalization.

We have informally tested HiGSFA on other problems (e.g., face detection, digit and traffic sign recognition) with good results. Therefore, we are interested in a formal and systematic evaluation of HiGSFA on these and more problems.

The idea of complementing GSFA with information preservation in hierarchical networks can also be applied to SFA (e.g., either using iSFA nodes or an HiGSFA network and a linear graph with samples ordered by time). The improved feature slowness of HiSFA over HSFA might be useful to improve simulations based on SFA for Neuroscience. Moreover, existing neural models based on the slowness principle might benefit from incorporating information preservation.

\subsection{Final words} 
We believe it is possible to develop successful learning algorithms based on a few simple but strong learning principles and heuristics, and this is the approach that we try to pursue with HiGSFA. We are not so much interested in developing an algorithm that might be strong but cannot be understood analytically.


The slow and reconstructive parts of the extracted features can be seen as two information channels, the first one codes information connected to the slow parameters, and the second one codes information representing the input. Although the information in the slow part is somewhat mixed, it can be further decomposed into three channels. The 3rd slowest feature is mostly related to race, the 4th one to gender, and the remaining ones to age.
Therefore, HiGSFA follows two of the suggestions in \cite{KruegerJanssenEtAl-2013} based on findings from Neuroscience on the primate visual system for successful computer vision, namely, hierarchical processing and information-channel separation. 


The proposed algorithm is general purpose (e.g., it does not know anything about face geometry), but it is still capable of outperforming special-purpose state-of-the-art algorithms, at least for the age estimation problem. 
This shows the improved versatility and robustness of the algorithm, and makes it a good candidate for many other problems of computer vision on high-dimensional data, particularly those lying at the intersection of image analysis, nonlinear feature extraction, and supervised learning.





\bibliography{references}

%

\end{document}